\documentclass{article}

\usepackage{PRIMEarxiv}

\usepackage[utf8]{inputenc} 
\usepackage[T1]{fontenc}    
\usepackage{hyperref}       
\usepackage{url}            
\usepackage{booktabs}       
\usepackage{amsfonts}       
\usepackage{nicefrac}       
\usepackage{microtype}      
\usepackage{lipsum}
\usepackage{fancyhdr}       
\usepackage{graphicx}       
\graphicspath{{media/}}     

\usepackage{microtype}
\usepackage{graphicx}
\usepackage{subfigure}
\usepackage{booktabs} 

\usepackage{hyperref}

\usepackage{algorithm}
\usepackage{algorithmic}

\newcommand{\alglinelabel}{%
  \label
}

\usepackage{amsmath,amssymb,comment, amsthm}
\usepackage{url}
\usepackage{graphicx}
\usepackage{subfigure}
\usepackage{color}
\usepackage{epsfig}
\usepackage{latexsym}
\usepackage{mathtools}
\usepackage{stackengine}
\usepackage{mathtools}
\usepackage{dsfont} 
\usepackage{makecell}
\usepackage{enumitem}
\usepackage{tabularx}
\usepackage{booktabs}
\usepackage{multicol}
\usepackage{bm}

\usepackage{natbib}

\newtheorem{theorem}{\bf Theorem}[section]

\newtheorem{definition}{Definition}[section]
\newtheorem{lemma}[theorem]{\bf Lemma}

\newtheorem{corollary}[theorem]{Corollary}

\newtheorem{remark}{Remark}

\newtheorem{assumption}{\bf Assumption}

\newcommand{\E}{\mathbb{E}}
\newcommand{\Ex}[1]{\mathbb{E}\left[ #1 \right] }

\newcommand{\bestarm}{i^{**}}

\ifodd 0
\newcommand{\dtl}[1]{{\color{magenta}Detail: #1}}
\else
\newcommand{\dtl}[1]{}
\fi

\ifodd 1
\newcommand{\com}[1]{{\color{red}Comment: #1}}
\else
\newcommand{\com}[1]{}
\fi

\newcommand{\ind}[1]{{\mathds{1}\left\{ #1\right\}}}

\mathtoolsset{showonlyrefs,showmanualtags}

\allowdisplaybreaks

\usepackage[textsize=tiny]{todonotes}

\title{Bandits with Anytime Knapsacks
}

\author{
  Eray Can Elumar \\
  Dept. of Electrical and Computer Engineering\\
  Carnegie Mellon University \\
  Pittsburgh, PA\\
 \texttt{eelumar@andrew.cmu.edu} \\
   \And
  Cem Tekin \\
  Dept. of Electrical Engineering\\
  Bilkent University \\
  Ankara, Turkey\\
  \texttt{cemtekin@ee.bilkent.edu.tr} \\
  \And 
  Osman Ya\u gan \\
  Dept. of Electrical and Computer Engineering\\
  Carnegie Mellon University \\
  Pittsburgh, PA\\
  \texttt{oyagan@andrew.cmu.edu} \\
}

\begin{document}
\maketitle

\begin{abstract}
We consider bandits with anytime knapsacks (BwAK), a novel version of the BwK problem where there is an \textit{anytime} cost constraint instead of a total cost budget. This problem setting introduces additional complexities as it mandates adherence to the constraint throughout the decision-making process. We propose SUAK (Strategic Under-utilization for Anytime Knapsacks), an algorithm that utilizes upper confidence bounds to identify the optimal mixture of arms while maintaining a balance between exploration and exploitation. SUAK is an adaptive algorithm that strategically utilizes the available budget in each round in the decision-making process and skips a round when it is possible to violate the anytime cost constraint. In particular, SUAK slightly under-utilizes the available cost budget to reduce the need for skipping rounds. We show that SUAK attains the same problem-dependent regret upper bound of $ O(K \log T)$ established in prior work under the simpler BwK framework, where $K$ denotes the number of arms, and $T$ denotes the time horizon. We also establish an instance-dependent lower bound showing that the $O(K\log T)$ dependence is unavoidable. Finally, we show via extensive simulations that SUAK achieves considerably higher rewards than naive adaptations of BwK algorithms that do not optimize learning under anytime cost constraints.
\end{abstract}

\section{Introduction}
Multi-armed bandits (MAB) is one of the fundamental problems in the field of sequential decision-making under uncertainty. In its essence, it is a problem setting where an agent must strategically allocate resources among the arms to maximize cumulative reward over time, navigating the trade-off between gathering information about uncertain arms (exploration) and exploiting known information to optimize immediate rewards (exploitation). This problem finds applications across diverse domains, including reinforcement learning \citep{intayoad2020reinforcement}, online advertising \citep{slivkins2013dynamic},  clinical trials \citep{villar2015multi}, and resource allocation \citep{soare2015sequential}.

The {\it Bandits with Knapsacks} (BwK) problem, introduced by \citet{Badanidiyuru2013BanditsWK}, is an extension of the classical multi-armed bandit problem, with the additional constraint of limited resource capacity akin to the {\em knapsack} problem \citep{tran2012knapsack}. In this scenario, an agent is confronted with a set of arms, each associated with an {\em unknown} reward and cost distribution. Unlike the traditional bandit setting, selecting an arm incurs both a reward and a cost here, and the objective is to maximize the total reward while respecting the total capacity constraint of the knapsack. The BwK problem encapsulates the trade-off between exploration and exploitation while managing resource constraints, presenting a rich framework with applications such as online advertising \citep{bwk_online_advertise, badanidiyuru2018bandits}, dynamic resource allocation \citep{kumar2022non_resource_alloc_bwk}, and personalized recommendations \citep{yu2016linear_bwk_recommendation_system}.

In this paper, we consider a specific variant of this problem, which we call the bandits with anytime knapsacks (BwAK) problem, where instead of a total cost budget there is an anytime constraint on the average cost. This problem setting introduces an additional level of complexity as a mixture strategy needs to be employed to be able to pull arms with mean costs higher than the average cost budget without violating the anytime constraint.

\subsection{Limitations of the State-of-the-Art} 

Standard BwK algorithms are designed for a \emph{total} resource budget over the horizon. A policy can satisfy a budget $cT$ even if its early consumption exceeds rate $c$, provided that its terminal consumption remains within the total budget. BwAK instead requires
\[
S_c(t)\leq ct,\qquad \forall t\geq 1,
\]
where $S_c(t)$ is the cumulative realized cost through round $t$. Hence a policy that is feasible under the standard BwK criterion can violate the BwAK constraint at an intermediate round. At round $t$, the learner can use only the accumulated budget slack $ct-S_c(t)$. Existing BwK algorithms use, among other approaches, optimistic estimates of the reward--cost optimization problem, primal--dual resource prices, or allocations based on the remaining budget and horizon; these methods are reviewed in \S\ref{sec:relatedwork}. None of these approaches, without an additional feasibility mechanism, guarantees the prefix-wise constraint above for every realized cost sequence \citep{Badanidiyuru2013BanditsWK,agrawal2014bandits,flajolet2015logarithmic_knap_2,li2021symmetry_knap_1}.

A natural adaptation to overcome this is to pull the arm selected by a BwK rule only when that pull cannot violate the current prefix constraint and to pull the null arm otherwise. This enforces feasibility, but the underlying BwK rule is not designed to accumulate budget slack for future high-cost pulls. It can therefore remain close to the constraint boundary, increasing the number of null-arm pulls imposed by this feasibility check. We introduce a baseline, \emph{Safe-Skip BwK (SS-BwK)}, in \S\ref{sec:naive_algorithm}; it applies this feasibility check to a standard BwK decision rule. The simulations show that this direct adaptation can incur more regret from null-arm pulls than a policy that considers the anytime constraint when deciding which arm to pull.

\subsection{On the Importance of Anytime Constraints} \label{sec:applications}

The formulation of the anytime constraint considered here has broad applications across various fields. A notable example is inventory management, where a factory produces goods at a constant rate and seeks to maximize revenue by selling to buyers in a marketplace, where bids consisting of price and order size are placed.
Our anytime constraint is especially relevant in such scenarios since having a negative inventory is not possible. 
An important aspect of this constraint is that it introduces the trade-off between exploiting the available inventory, and skipping a round to accumulate more inventory in order to capture bids with higher order sizes. 
This highlights the added complexity of our problem setting and underscores its broader applicability in settings beyond the standard BwK framework.

Another example is in clinical trials, where drugs or treatment plans with unknown costs, which can be side effects; and rewards, which can model treatment efficacy; need to be evaluated. This example can be modeled using the anytime constraint if participants are recruited to the trials at a constant rate per week, and are not allowed to trial more than one drug. In online advertising, an advertiser may have a daily budget limit to prevent overspending. In this context, the `arms' represent different ad campaigns or strategies, each with varying costs that can be selected for the day. The reward can be modeled as the daily revenue generated from clicks or the number of users who subscribe.
One last example is in satellite systems, where solar panels generate energy and excess energy can be stored in a battery. Here, $c$ can correspond to the energy generated per unit time, and arms can correspond to different tasks that need to be performed, with their rewards reflecting the importance of the tasks. The cost of an arm then represents the energy needed for the task.

\subsection{Contributions}

 \begin{enumerate}[leftmargin=*]

  \item \textbf{SUAK Algorithm:} SUAK utilizes the upper confidence bounds to explore the optimal base that solves the
problem, and also strategically under-utilizes the available budget to limit the number of rounds that are skipped when satisfying the anytime constraint. It is also horizon-free; SUAK takes only the average cost target $c$ as an input parameter, and does not need knowing $T$ beforehand.
\item \textbf{Theoretical Analysis:} We provide an $O(K\log T)$ instance-dependent upper bound on the regret of SUAK which shows linear dependence on the number of arms. We also show an instance-dependent regret bound of $\Omega(K\log T/\Delta)$ where $\Delta$ is the reward gap of an arm.
 \item \textbf{Technical Novelty:} Our analysis departs from classical UCB-style proofs by exploiting the control logic of
SUAK, which pulls an arm whenever its cost confidence interval lies inside $c$ and skips rounds to enforce
anytime cost constraints. This structure enables a nonstandard conditioning as the estimation error on the arm cost
gap will always be smaller than a threshold value. This also implies a lower bound on effective sample size for
arms, yielding tight multiplicative bounds, while a pull-probability lower bound on an arm couples base selections
to arm-level sampling without over-counting. Further, We jointly bound the regret due to under-budgeting and global skips by analyzing a coupled process with a negative-drift argument above a suitable barrier and Doob’s optional stopping theorem applied to a localized exponential supermartingale.

\end{enumerate}

The remainder of the paper is organized as follows. In \S\ref{sec:prob_state}, we formally describe the problem setting, the optimal strategy, and the notion of regret. Related work is provided in \S\ref{sec:relatedwork}. In \S\ref{sec:algorithm_general}, we first present the naive UCB-based baseline in \S\ref{sec:naive_algorithm} and then introduce \textsc{SUAK} in \S\ref{sec:algorithm}. Theoretical guarantees for \textsc{SUAK}, including the proof of the main theorem, are provided in \S\ref{sec:analysis}. Simulation results for \textsc{SUAK} are provided in \S\ref{sec:simulations}, followed by concluding remarks in \S\ref{sec:concl}. Proofs of auxiliary results are deferred to the Appendix.

\section{Problem Statement}
\label{sec:prob_state}

We consider a $K$-armed stochastic bandit problem with the set of non-null arms $[K]$. For each arm $i\in[K]$, the reward--cost pairs $\{(r_i(t),\rho_i(t))\}_{t\ge1}$ are i.i.d. across rounds under a fixed joint distribution $P_i$ supported on $[0,1]^2$, with
\begin{align}
    \E_{P_i}[r_i(t)]=\mu_i,
    \qquad
    \E_{P_i}[\rho_i(t)]=\rho_i.
\end{align}
The joint processes are independent across arms. We allow arbitrary dependence between reward and cost within the same arm and round. At each round $t$, the agent has the option of skipping by not pulling any of the $K$ arms. We model this decision by introducing an arm with cost and reward equal to $0$, denoted by $\bot$, which is known as the {\it null arm} in the BwK literature. We let $\mathcal A := [K] \cup \{\bot\}$ denote the action set, and set $\rho_{\bot}=\mu_{\bot}=0$. Accordingly, $\boldsymbol{\rho} = [\rho _1,\cdots,\rho _K, \rho_{\bot}]^T $ and $\boldsymbol{\mu} = [\mu_1,\cdots,\mu_K, \mu_{\bot}]^T $ denote the mean cost vector and the mean reward vector indexed by $\mathcal A$, respectively. Throughout this paper, we use bold symbols to denote vectors or matrices. We use $\Delta_m:=\{x\in\mathbb R_+^m:\mathbf 1^\top x=1\}$ for the probability simplex with $m$ coordinates. We let $\bestarm := \arg \max_{i \in [K]} \mu_i$ denote the arm with the highest mean reward and, for simplicity, assume this arm is unique.

Let $i(t)$ be the arm pulled by the agent in round $t$, $r(t)$ represent the reward received in round $t$, and $c(t)$ represent the cost incurred in round $t$. Let \((\mathcal F_t)_{t\ge0}\) denote the filtration generated by the observed history, where \(\mathcal F_{t-1}\) contains all information available before round \(t\). The i.i.d. assumption above implies that, for every $i\in\mathcal A$,
\begin{align}
 \mathcal L\!\left((r(t),c(t))\mid \mathcal F_{t-1},\,i(t)=i\right)&=P_i, \label{eq:joint_freshness}\\
 \E[r(t)\mid \mathcal F_{t-1},\,i(t)=i]&=\mu_i,\qquad
 \E[c(t)\mid \mathcal F_{t-1},\,i(t)=i]=\rho_i,
 \label{eq:conditional_means_freshness}
\end{align}
In particular, adaptively selecting an arm using the preceding observed history does not alter the distribution of its current reward and cost observations. Note that the null arm is deterministic $(r_\bot(t),c_\bot(t))=(0,0)$. To make all action-selection quantities predictable, let
\begin{align}
N_i(t):=\sum_{s=1}^{t-1}\ind{i(s)=i}
\end{align}
denote the number of times arm $i$ is pulled immediately before round $t$. Define $S_c(t) = \sum_{s=1}^t c(s)$ and $\Bar{c}(t) = S_c(t)/t $ as the cumulative cost and the average cost incurred through round $t$. Whenever $N_i(t)>0$, we define the empirical reward and cost means as
\begin{align}
\Bar{\rho}_i(t)
&:=\frac{\sum_{s=1}^{t-1}\rho_i(s)\ind{i(s)=i}}{N_i(t)},\qquad
\Bar{\mu}_i(t)
:=\frac{\sum_{s=1}^{t-1}r_i(s)\ind{i(s)=i}}{N_i(t)}.
\end{align}
 Define the realized cumulative reward through round $t$ by $F(t):=\sum_{s=1}^t r(s)$. We assume that there is an average cost budget of $c$ per round that cannot be exceeded at any round, which we refer to as the anytime cost constraint. The agent aims to maximize the expected cumulative reward received under this constraint. This objective can be
expressed as:
\begin{align}
    \max\ \E[F(t)] \qquad \text{s.t.}\qquad \frac{\sum_{s=1}^u c(s)}{u} \leq c, \quad \forall u\le t.
\end{align}

We call this setting, which captures many practical applications as discussed in \S~\ref{sec:applications}, the {\em Bandits with Anytime Knapsacks (BwAK)} problem. Ideally, regret would be measured relative to the optimal policy that knows all reward and cost distributions and maximizes expected reward while satisfying the anytime constraint. Because this policy is difficult to characterize, we instead use the following linear relaxation and define regret relative to its optimal value.

\textbf{Linear Relaxation.} Following the prior work, we consider the following linear relaxation: 

\begin{align}
    OPT_{LP}(T) =  \max_{\boldsymbol{\pi}} \ & T \cdot \boldsymbol{\mu}^T \boldsymbol{\pi} \label{def:opt_lp} \\   \text{ s.t.  } & \boldsymbol{\rho}^T \boldsymbol{\pi} \leq c, \  \boldsymbol{\pi} \in \Delta_{K+1}.   
\end{align}
where the vector $\boldsymbol{\pi}$ represents the policy which defines the fraction of time an arm will be pulled. Because the feasible region is a polytope defined by a single resource constraint and the simplex boundary, standard linear programming theory implies that an optimal basic feasible solution exists with at most two non-zero coordinates (i.e., at most two arms will have nonzero $\pi_i$ values). We refer to the support of a basic feasible solution as a \emph{base} and, use $\mathds V$ exclusively for the canonical feasible LP vertices:
\begin{align}
\mathds V
:=&\ \bigl\{\{i\}:i\in\mathcal A,\ \rho_i\le c\bigr\} \\
&\ \cup\bigl\{(h,\ell):h,\ell\in\mathcal A,\ h\ne\ell,\ \rho_h>c>\rho_\ell\bigr\}.
\label{eq:canonical_vertex_set}
\end{align}
Thus a two-arm base is a cost-$c$ boundary vertex with its higher-cost arm listed first, while a one-arm base is a feasible single arm. Note that $\{\bot\}\in\mathds V$, and a support consisting of two arms whose mean costs are both strictly below $c$ cannot be considered a two-arm base.

We let $\mathcal{I}^i := \{\mathcal{I} \in \mathds{V}: i \in \mathcal{I}\}$ denote the set of bases that include arm $i$, let $\boldsymbol{\pi}^*$ denote an optimal basic feasible solution of \eqref{def:opt_lp}, and let $r^* := \boldsymbol{\mu}^T\boldsymbol{\pi}^*$ be its optimal reward per round. We also define $\mathcal I^*:=\{i:\pi_i^*>0\}\in\mathds V$ as the optimal base. 

Let $OPT$ denote the total expected reward of a dynamic policy in $T$ rounds that conforms to a total budget constraint over $T$ rounds as in the standard BwK literature instead of the anytime budget constraint we consider here. It was shown that $OPT_{LP} \geq OPT$ \citep{Badanidiyuru2013BanditsWK}.

\setlength{\textfloatsep}{9pt}

Let $REF$ denote the total expected reward of a dynamic policy over $T$ rounds that conforms to the average cost constraint. This constraint is stricter than the total budget constraint. This can easily be seen as satisfying the anytime constraint in the last round $T$ with $c=B/T$ produces the total budget constraint of $B$ over $T$ rounds. Hence, $OPT_{LP} \geq OPT \geq REF$. While regret could be defined as the difference between expected cumulative reward of SUAK and $REF$, we choose a stronger regret definition by defining it with respect to $OPT_{LP}$ as 
\begin{align}
   R_T = OPT_{LP} - \Ex{F(T)} = T r^* - \Ex{F(T)} \label{eq:def_regret}
\end{align}
so that our results can be compared with prior work on the total budget setting. 
Notation for terms used throughout the paper is provided in Table \ref{table:notations}.

\begin{table}[h]
    \caption{Notations}
    \begin{tabularx}{\textwidth}{p{0.22\textwidth}X}
    \toprule
      $K$ & number of non-null arms  \\
      $\mathcal A=[K]\cup\{\bot\}$ & action set, i.e. the set of all arms including the {\it null arm} $\bot$ \\
      $\boldsymbol{\rho} $ & mean cost vector indexed by $\mathcal A$ \\
      $\boldsymbol{\mu} $ & mean reward vector indexed by $\mathcal A$ \\
      $\mathds{V}$ & set of canonical feasible LP vertices (bases), as defined in \eqref{eq:canonical_vertex_set} \\
      $\rho_i(t)$ & realized cost of arm $i$ in round $t$  \\
      $c(t)=\rho_{i(t)}(t)$ & cost incurred in round $t$ \\
      $\rho_i$ & mean cost of arm $i$   \\
      $\Bar{\rho}_i(t)$ & empirical mean cost of arm $i$ before round $t$  \\ 
      $\epsilon_i(t)$ & $\epsilon_i(t) = \sqrt{\frac{3 \log t}{N_i(t)}}$, the confidence radius of arm $i$ before round $t$  \\      
      $\rho^L_i(t)$ & $\rho^L_i(t)=\operatorname{proj}_{[0,1]}\!\left(\Bar{\rho}_i(t)-\epsilon_i(t)\right)$, the clipped cost LCB before round $t$  \\
      $\mu^U_i(t)$ & $\mu^U_i(t)=\operatorname{proj}_{[0,1]}\!\left(\Bar{\mu}_i(t)+\epsilon_i(t)\right)$, the clipped reward UCB before round $t$  \\      
   $\varrho^L_l(t)$ &  $\varrho^L_l(t) := \Bar{\rho}_l(t) - 7\sqrt{1.5 \log t / N_l(t)}$  \\ 
   $\varrho^U_l(t) $ & $ \varrho^U_l(t):= \Bar{\rho}_l(t) + 7\sqrt{1.5 \log t / N_l(t)}$ \\ 
    $\omega(t)$ & $\omega(t):= \delta_{\min}^L(t) / (2+\delta_{\min}^L(t)-c)$, the under-budgeting amount at round $t$ \\
$\delta_{\min}^L(t)$ & $\delta_{\min}^L(t) := \min_{i\in[K]} (| \bar{\rho}_i(t) - c| - \sqrt{1.5 \log t / N_i(t)})$, LCB of minimum cost gap \\
      $r_i(t)$ & reward obtained from pulling arm $i$ in round $t$  \\ 
      $r_{\mathcal{I}}$ & reward of base $\mathcal{I}$  \\ 
      $\Delta_{\mathcal I}$ & $\Delta_{\mathcal I}:=r^*-r_{\mathcal I}$, the reward gap of base $\mathcal I$  \\ 
      $\mu_i$ & mean reward of arm $i$   \\    $\boldsymbol{\pi}^* $ & optimal solution of the problem under the linear relaxation \\
      $\bestarm $ &  $\bestarm := \arg \max_{i \in [K]} \mu_i$ \\
      $S_c(t)$ & $ S_c(t) := \sum_{s=1}^t c(s)$ cumulative cost incurred until round $t$  \\ 
      $\Bar{c}(t) $ & $\Bar{c}(t) = S_c(t)/t $ average cost incurred at round $t$ \\
        $N_i(t)$ & number of times arm $i$ has been pulled before round $t$\\
        $N_{(i,j)}(t)$ & number of times base $(i,j)$ has been selected before round $t$\\
        $\Delta_i $ &  $\Delta_i := \mu_{\bestarm} - \mu_i$, the reward gap of an arm \\
        $\delta_i $ &  $\delta_i := |\rho_i - c|$, the cost gap of an arm \\
        $\delta_{\min} $ &  $\delta_{\min} := \min_i \delta_i$, the minimum cost gap \\
        $g_i$ & $g_i:=\nu^*+\lambda^*\rho_i-\mu_i$, the reduced-cost gap of arm $i$ \\
        $G_{\mathcal I}$ & structural charge: $\max_{q\in\mathcal I}g_q$ for a two-arm base and $\Delta_{\mathcal I}$ for a singleton \\
        $\lambda^*$ & optimal dual multiplier (shadow price) of the average-cost constraint \\
        $\Delta_{\min,i} $ & $\Delta_{\min,i}:=\min\{\Delta_{\mathcal I}:\mathcal I\in\mathds V,\ i\in\mathcal I,\ \Delta_{\mathcal I}>0\}$, with $\Delta_{\min,i}=\infty$ if no suboptimal base contains $i$ \\
      \bottomrule
     \end{tabularx}
     \label{table:notations}
    \end{table}

\subsection{Related Works}
\label{sec:relatedwork}

In this subsection, we present some of the works that are relevant to our problem setting. 

 \textbf{Bandits with knapsacks:} 
 The BwK problem has been studied before and algorithms that achieve optimal problem-independent regret bounds of $O(\sqrt{K \cdot OPT})$ have already been developed \citep{ agrawal2014bandits}. However, deriving problem-dependent lower bounds that reflect the resource-learning difficulty, and developing algorithms that attain such bounds, are still open questions.
One notable work in this field is by
\citet{flajolet2015logarithmic_knap_2}, in which the BwK problem is considered under three cases with $1$, $2$; and $d$ constraints. For $2$ constraints, 
a regret bound of $O\left( \lambda^2 K^2 \log T/ (\delta_{\min}^3 \Delta )  +  K^2 \sigma \log T /\delta_{\min}^3\right)$ is achieved where $\delta_{\min}$ is the minimum cost gap; $\sigma$ is the minimum $1/ \mu$ value where $\mu$ denotes the mean reward; $\lambda = 1 + 2 \kappa$; and $\kappa$ is a constant assumed to be known  {\em a priori} such that $|\mu_i-\mu_j|\leq \kappa |\rho_i - \rho_j|$ for any $i,j$. For the $d$ constraint setting, a regret bound of $O(2^{K+d} \log T )$ is achieved. The $2$-constraint setting is similar to our work, as our one-dimensional cost, and the anytime cost constraint add up to two constraints for the LP. Their regret bound has quadratic dependence on $K$, whereas our regret bound is linear in $K$. The lower bound in \S~\ref{sec:lower_bound} shows that linear dependence on $K$ together with logarithmic dependence on $T$ is unavoidable on some problem instances.

Another notable prior work is by \citet{li2021symmetry_knap_1}, where a $d$-dimensional cost vector is considered. 
They propose a two-phase algorithm where the first phase pulls each arm the same number of times until all the suboptimal arms are eliminated. In the second phase, the base with the highest UCB index is chosen. Their analysis also develops LP-based decompositions of regret and uses remaining-resource LP structure; these are established tools that our proof adapts to the anytime setting rather than claims as new. 
They achieve a regret bound of $O(Kd \log T / (b^3 \Delta^2) + d^4/(b^2 \min \{\chi^2, \Delta^2\} \min \{1, \sigma^2\}))$, where $\Delta$ in their setting is defined as the gap between the reward of the optimal solution per round and the maximum reward that can be obtained per round when one arm is removed; $b$ is the average cost budget per round; $\chi$ is the minimum nonzero value in the optimal policy; $\sigma$ is a constant related to the linear dependency between arms. 
\begin{table*}[t]
  \caption{Comparison of our work with prior work on bandits with knapsacks }
  \label{priorwork-table}
  \centering
  \begin{tabular}{lllll}
    \toprule
    Work     & Model     & Regret Bound \\
    \midrule
    \textbf{Our work} & \makecell[l]{Shared stochastic cost; hard prefix budget}  & $O\!\left(\frac{K\log T}{\delta_{\min}^3\Delta_{\min}}+\frac{\nu^*K\log T}{c\delta_{\min}^3}+\frac{K\log T}{c\delta_{\min}^2}\right)+O(1)$ \\
\citet{combes2015bandits}& \makecell[l]{Arm-specific evolving budgets} & $O(\log T)$ \\
\citet{bernasconi2024beyond}& \makecell[l]{BwRK with sublinear violations}  & $  O \left( \sqrt{T} \right)$ \\
\citet{kumar2022non_resource_alloc_bwk} & \makecell[l]{Non-monotonic resource drift}  & \makecell[l]{$\widetilde C\log T$, where $\widetilde C$ includes\\$\frac{Km^2}{\Delta^2\min\{\delta_{\rm drift}^2,\sigma_{\min}^2\}}$, $K(\gamma^*)^{-2}$, and control terms} \\     
\citet{li2021symmetry_knap_1}  & \makecell[l]{ Total  budget}  & $O\left( \frac{K d \log T}{b^3 \Delta^2}+\frac{d^4}{b^2 \min \left\{\chi^2, \Delta^2\right\} \min \left\{1, \sigma^2\right\}}\right)$    \\
\citet{flajolet2015logarithmic_knap_2}    & \makecell[l]{ Total  budget} & $ O\left( \frac{\lambda^2 K^2 \log T}{\delta_{\min}^3 \Delta}  + \frac{K^2 \sigma \log T}{\delta_{\min}^3}\right)$       \\
    \bottomrule \label{table:priorwork}     
  \end{tabular}
\end{table*}

\textbf{Bandits with evolving arm-specific budgets:} \citet{combes2015bandits} study stochastic bandits in which each arm has its own availability budget. Their general-budget formulation permits these arm-specific budgets to increase over time and even to depend on the sample path. They introduce the B-KL-UCB algorithm which attains $O(\log T)$ regret. However, this setting differs structurally from BwAK. Exhausting the budget of one arm only changes the availability of that arm, and their oracle policy greedily selects the highest-mean available arm. In BwAK, by contrast, all arms consume the same cost budget and the cumulative incurred cost must satisfy $\sum_{s=1}^t c(s)\le ct$ for every $t$. A high-cost arm can therefore be used only by balancing it with lower-cost or null actions, and the LP optimum may be a boundary mixture of arms on opposite sides of $c$. Hence, while their logarithmic result is important for evolving budgets, it does not directly cover the BwAK problem we consider here.

\textbf{Bandits with Replenishable Knapsacks: } In this setting, cost of an arm is allowed to be negative, which allows the knapsack to be replenished. In \citet{bernasconi2024no}, a more general BwK formulation with long-term constraints is considered, which stipulates the total consumption of each resource at round $T$ should be less than zero up to small sublinear violations. The costs can be negative as well as positive. This is again similar to our setting as our setting can be viewed as their setting where sublinear violations are not allowed. They propose a Primal-Dual algorithm-based framework that uses the {\emph EXP3-SIX} algorithm, and achieve $O(\sqrt{KT}\log(KT))$ instance-independent regret bound. They also remark that initial $o(T)$ rounds can be skipped to cover the potential violations and implement the long-term constraint as a hard constraint like in our setting. They provide an upper bound of $O(\sqrt{KT})$ constraint violations in \citet[Corollary 8.2]{bernasconi2024no}, which suggests an upper bound of $O(\sqrt{KT})$ on the number of initial rounds that would need to be skipped to achieve hard constraints. Note that this upper bound might not be tight. This would lead to $O(\sqrt{KT})$ gap-independent regret in our problem setting. In our work, we focus on instance-dependent regret bound and show that we achieve $O(K \log T)$ instance-dependent regret for the same problem setting. In \citet{bernasconi2024beyond}, they improve their prior work \cite{bernasconi2024no} by using a UCB-based approach which lets them provide $O(\sqrt{T})$ regret in stochastic settings without assuming Slater's condition. The upper bound on constraint violations is still $O(\sqrt{KT})$, which would lead to an upper bound of $O(\sqrt{KT})$ for the instance-independent regret in our problem setting.  Another notable prior work is by \citet{slivkins2024contextualbanditspackingcovering}, 
where packing and covering constraints, as well as negative resource consumption, are considered. Their algorithm works when the initial budget is $B=\Omega(T)$, or $B=o(T)$, compared to the prior work which mostly restricts $B$ to $B=\Omega(T)$. This is similar to our setting as our setting can be reduced to theirs by implementing the budget increase of $c$ by subtracting $c$ from the costs of all arms including the null arm. However, their algorithm is suboptimal in our setting with $B=0$, as they remark in the discussion of \citet[Theorem~3.6]{slivkins2024contextualbanditspackingcovering}, that their proposed algorithm LagrangeCBwLC 
is suboptimal when $B= o(T)$.
This is as expected since Lagrange-based algorithms generally require knowing the ratio $T/B$, which goes to infinity when $B= o(T)$. 

In \citet{bernasconi2024bandits_replenishable}, there exists an arm with a negative expected cost that allows to replenish the budget. However, this work cannot be used in our setting as their work can only be used when $B=\Omega(T)$,  
and they only provide instance-independent bounds of $O(\sqrt{KT})$, whereas we provide $O(K\log T)$ instance-dependent bounds.

\textbf{BwK with non-monotonic resource utilization:} This setting generalizes the standard BwK model by allowing each action to induce a resource-drift vector whose components may be positive, negative, or zero. In each round, the learner observes both the reward and the realized resource drift, and the available budget of each resource is updated by the corresponding drift. \citet{kumar2022non_resource_alloc_bwk} allow actions to have positive, negative, or zero expected resource drift and establish logarithmic regret with respect to an LP benchmark under nondegeneracy and separation assumptions.
Their model admits a direct reduction to the single-resource BwAK setting considered in this paper. For each non-null arm \(i\), set its resource drift to
$d_i=c-\rho_i,$
set the null-arm drift to
$d_{\bot}=c,$
and initialize the resource budget at zero. Then, after \(t\) rounds, the accumulated resource budget is
$\sum_{s=1}^t d_{i(s)}
=
ct-\sum_{s=1}^t c(s).$
Therefore, requiring this budget to remain nonnegative at every round is exactly equivalent to the BwAK anytime constraint. Under the same reduction, their LP drift constraint coincides with \eqref{def:opt_lp}. Consequently, their results imply logarithmic instance-dependent regret for BwAK instances satisfying their assumptions. A direct quantitative comparison with SUAK is nevertheless difficult because the two analyses are expressed in terms of different problem-dependent parameters.
Under this reduction, their minimum drift-magnitude parameter becomes
\begin{align}
 \delta_{\mathrm{drift}}^{\rm KK}=\min\{c,\delta_{\min}\}.
\end{align}
If the unique optimal solution assumed in their analysis is a binding two-arm base $\{h,\ell\}$ with $\rho_h>c>\rho_\ell$, their support-separation parameter is
\begin{align}
 \delta_{\mathrm{support}}^{\rm KK}
 =\min\{\pi_h^*,\pi_\ell^*\}
 =\frac{\min\{\delta_h,\delta_\ell\}}{\delta_h+\delta_\ell},
\end{align}
and their conditioning parameter $\sigma_{\min}^{\rm KK}$ is the smallest singular value of the corresponding constraint matrix
\begin{align}
 D_{h\ell}=\begin{pmatrix}-\delta_h&\delta_\ell\\1&1\end{pmatrix}.
\end{align}
If, instead, the single resource constraint is nonbinding at a unique optimal singleton \(i^*\), their slack parameter becomes
$\delta_{\mathrm{slack}}^{\rm KK}=c-\rho_{i^*}=\delta_{i^*}$. These quantities enter their control margin and learning constant. They are conceptually different from SUAK's adaptive, data-dependent under-budgeting parameter \(\omega(t)\) defined in \eqref{eq:omega_t}.

The comparison can therefore be summarized more narrowly. Their algorithm uses the horizon \(T\), assumes a unique optimal LP support, and its regret bound has squared inverse dependence on its LP gap \(\Delta\). In contrast, SUAK is horizon-free: it takes \(c\), but not \(T\), as an input; it does not require uniqueness of the true optimal support or externally supplied gap-dependent control parameters; and its regret bound has inverse dependence on the base reward gap. These distinctions, however, do not imply uniform dominance of one approach over the other. SUAK assumes that every non-null arm is separated from the threshold \(c\), and its classification and reserve terms can become large when some \(\delta_i\) is small, even if that arm is not part of the optimal base.

\textbf{Conservative and safe bandits:} {\em Conservative bandits} is a framework where an action is considered safe at round $t$ if it keeps the cumulative reward up to round $t$ above a given fraction of a baseline policy. In \citet{wu2016conservative}, a baseline safe arm is given, and they provide an algorithm that utilizes the UCB principle where the baseline arm is pulled if the arm chosen by the UCB principle is not safe. In \citet{moradipari2020stage_conservative}, an anytime constraint is used instead of a constraint on cumulative reward that demands the expected reward of the pulled arm to be greater than a given threshold with high probability. They provide an algorithm that starts with a safe baseline and utilizes confidence regions to explore the other safe arms. {\em Safe bandits} is a similar framework to conservative bandits where in each round, the agent is required to select an arm with a given property no less than a predetermined (safe) threshold with high probability. For example, in \citet{hou2023probably_anytime_safe}, the agent needs to choose at most $K$ items from a set of $L$ items; and with probability at least $1-\delta$, the sum of variances of the selected items should not exceed a given threshold, which they call as the anytime-safe constraint. They propose a two step approach where first a set of arms is selected according to the UCB principle, and if this selection exceeds the threshold, the selected arms are split and pulled over multiple rounds. Our work is also similar in the sense that we also implement constraint checking as an additional separate step in SUAK.

A comparison between our work and prior literature is summarized in Table~\ref{table:priorwork}. Throughout this paper, $\Delta$ denotes the reward gap of an arm, while $\delta$ denotes the cost gap; we further define $\delta_{\min}$ as the minimum cost gap across all arms. These terms are formally introduced in \S~\ref{sec:analysis}. Due to differences in gap definitions and problem-dependent parameters, these results are not directly comparable. In \S~\ref{sec:lower_bound}, we derive such a lower bound by restricting BwAK to a deterministic-cost subclass that reduces exactly to the classical stochastic multi-armed bandit. This lower bound establishes that the $K\log T$ dependence of our upper bound is unavoidable.

\begin{algorithm}[tb]
\caption{Safe-Skip BwK (SS-BwK)}
\label{alg:naive_approach}
\begin{algorithmic}[1]
 \STATE \textbf{Input: } Avg. cost target $c$, number of rounds $T$
\STATE \textbf{Initialize: } Sample each arm once while skipping accordingly so that $\forall t\leq t_{\text{init}}$, $ S_c(t-1) + 1 \leq c\cdot t $ \label{alg_line:naive_init}
\FOR{each round $t \geq 1$ }
\IF{$S_c(t-1) + 1 > c\cdot t $} \alglinelabel{alg_line:naive_skip1} \alglinelabel{line_algnaive_1}
\STATE Skip the round 
\ELSE \alglinelabel{line_algnaive_2}
\STATE Let $B_r(t)=cT-S_c(t-1)$ and $H_r(t)=T-t+1$
\STATE 
$\begin{array}{c}
\text{Solve the following LP: } \quad \tilde{\boldsymbol{x}}(t)\in\arg\max_{\boldsymbol{x}\geq\mathbf{0}}\quad
    (\boldsymbol{\mu}^U(t))^{\top}\boldsymbol{x} \\
\quad\quad \hspace{2cm} \text{s.t.}\quad
    (\boldsymbol{\rho}^L(t))^{\top}\boldsymbol{x}\leq B_r(t),\\
    \hspace{2cm} \quad\quad \mathbf{1}^{\top}\boldsymbol{x}=H_r(t).
\end{array}$ 
\alglinelabel{alg_line:naive_lp}
\STATE Randomly pull arm $i$ with probability $\tilde{x}_i(t)/H_r(t)$
\ENDIF
\STATE Update $S_c(t)$ and the empirical statistics used to form $\boldsymbol{\rho}^L(t+1)$ and $\boldsymbol{\mu}^U(t+1)$
\ENDFOR 
\end{algorithmic} 
\end{algorithm}

\section{Algorithms for the BwAK}
\label{sec:algorithm_general}

Before presenting the SUAK Algorithm, to demonstrate the additional complexities of our problem formulation over the standard BwK setting, and also to serve as a baseline, we present a naive approach which makes it possible to convert any BwK algorithm to our BwAK setting.

\subsection{The Naive Approach}
\label{sec:naive_algorithm}

In this subsection, we consider a naive strategy that incorporates skipping into existing BwK algorithms, and show that it can lead to large regret. In this approach, in a round $t$, we first check if it is possible to violate the anytime constraint, and skip the round if it is the case. Otherwise, we let the BwK algorithm pull an arm. For this end, we use the {\em One Phase} Algorithm in \citet{li2021symmetry_knap_1}, and add skipping such that round $t$ is skipped if $S_c(t-1)+1>ct$. This implementation, which we call the {\em Safe-Skip BwK (SS-BwK)} Algorithm, is provided in Algorithm \ref{alg:naive_approach}.

In this algorithm, the initialization phase consists of sampling each arm once while using skips to prevent violation of the constraint. After this phase, we utilize a skipping mechanism in lines \ref{line_algnaive_1}--\ref{line_algnaive_2}, and if the round is not skipped, the algorithm proceeds to the linear program in line \ref{alg_line:naive_lp}. Here $\boldsymbol{\mu}^U(t)$ is the UCB of arm reward at round $t$ and $\boldsymbol{\rho}^L(t)$ is the LCB of arm cost at round $t$. We write $B_r(t)=cT-S_c(t-1)$ for the total budget remaining before round $t$ and $H_r(t)=T-t+1$ for the number of remaining rounds. The decision variable $\boldsymbol{x}$ is therefore an allocation-count vector over the remaining horizon, not a probability vector. In addition to the remaining-budget constraint, the LP explicitly includes the time-resource constraint $\mathbf{1}^{\top}\boldsymbol{x}=H_r(t)$. This equality is without loss of generality because the null arm has zero reward and zero cost and can absorb any otherwise unused allocation. Consequently, $\tilde{x}_i(t)/H_r(t)$ is already a probability distribution over arms, and SS-BwK samples the round-$t$ arm from this distribution.

\begin{figure*}
\centering
 \includegraphics[scale=0.15]{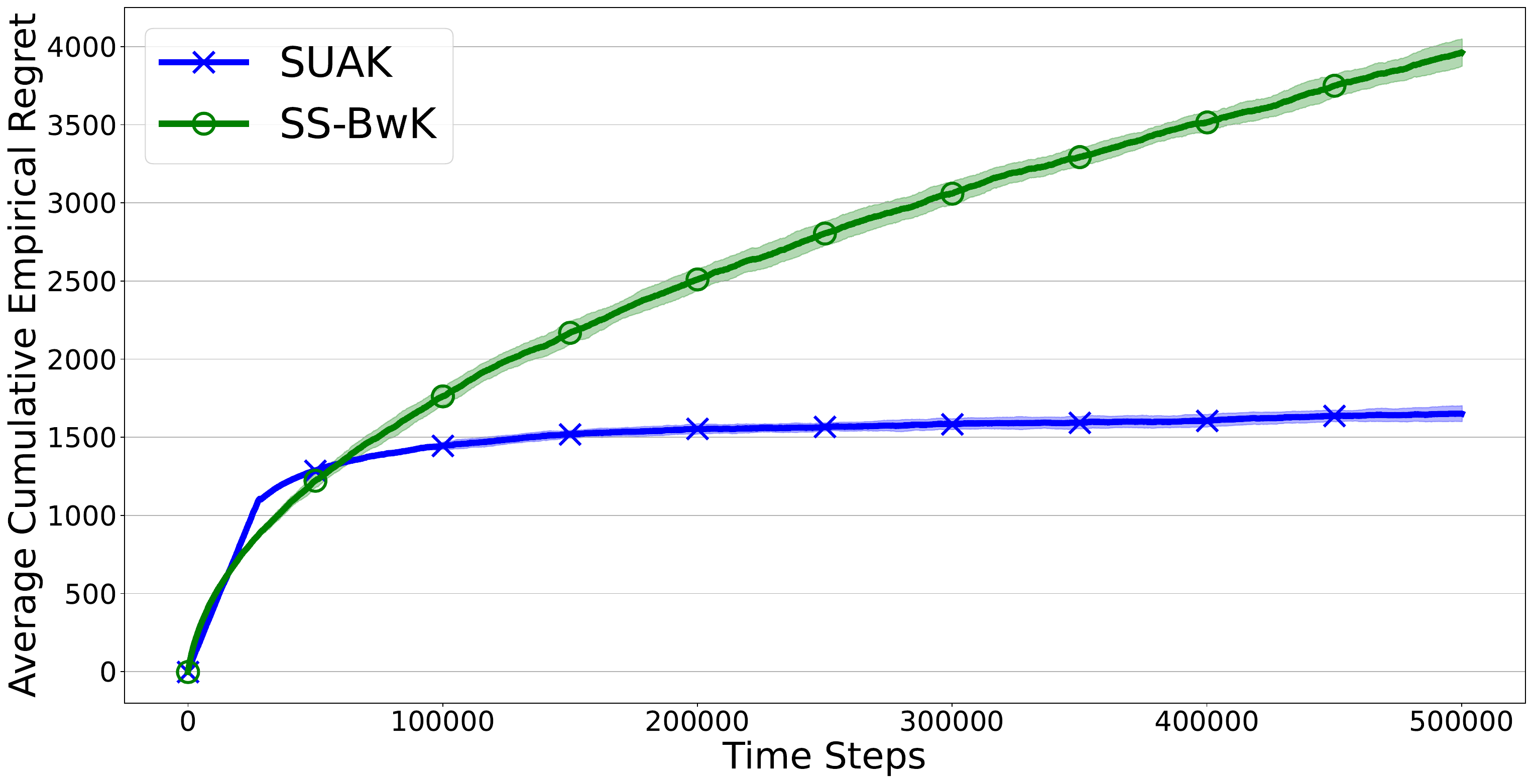} \hspace{3pt} 
\includegraphics[scale=0.15]{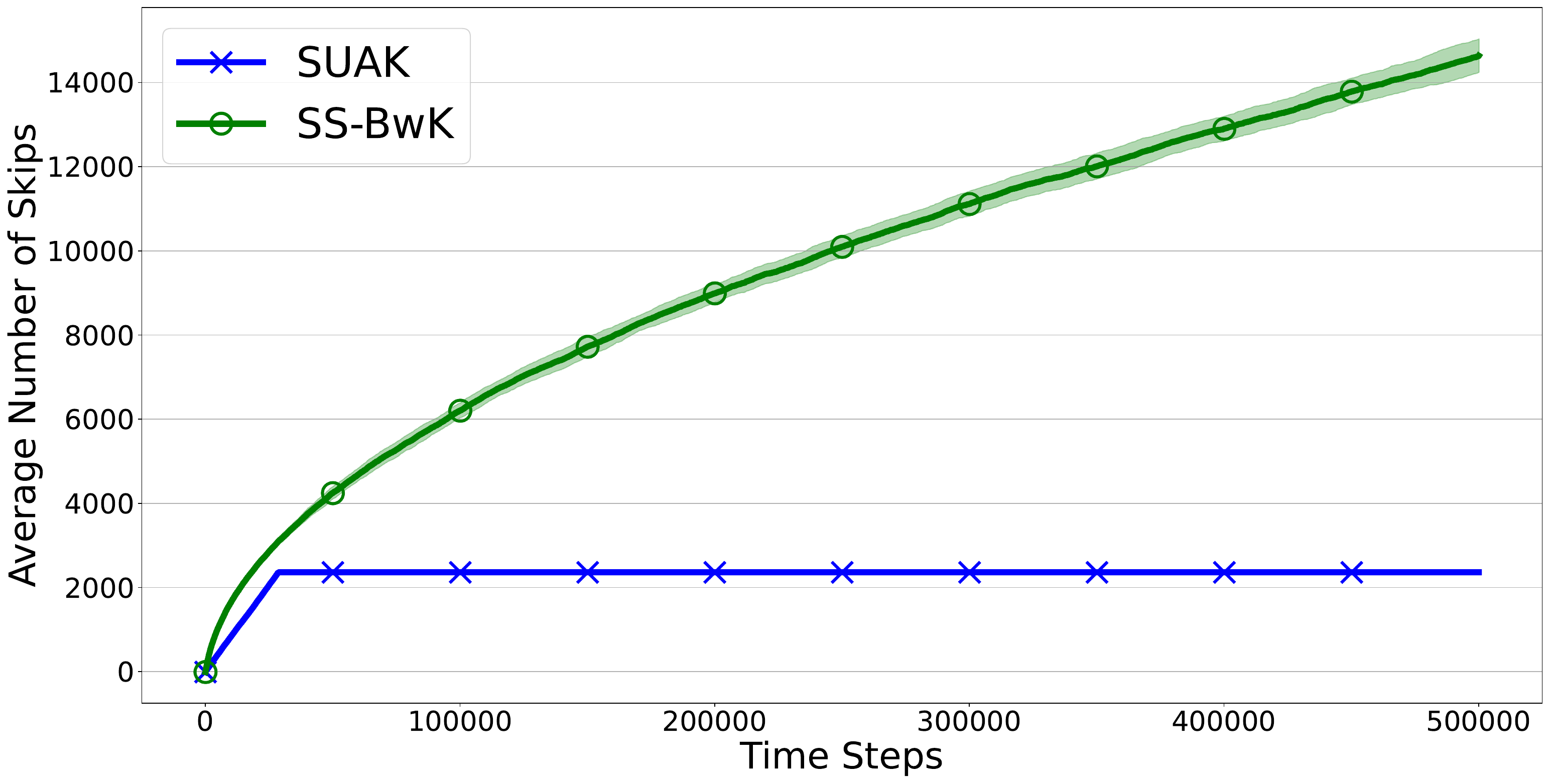} \\ \vspace{4mm} \hspace{3pt}
\includegraphics[scale=0.15]{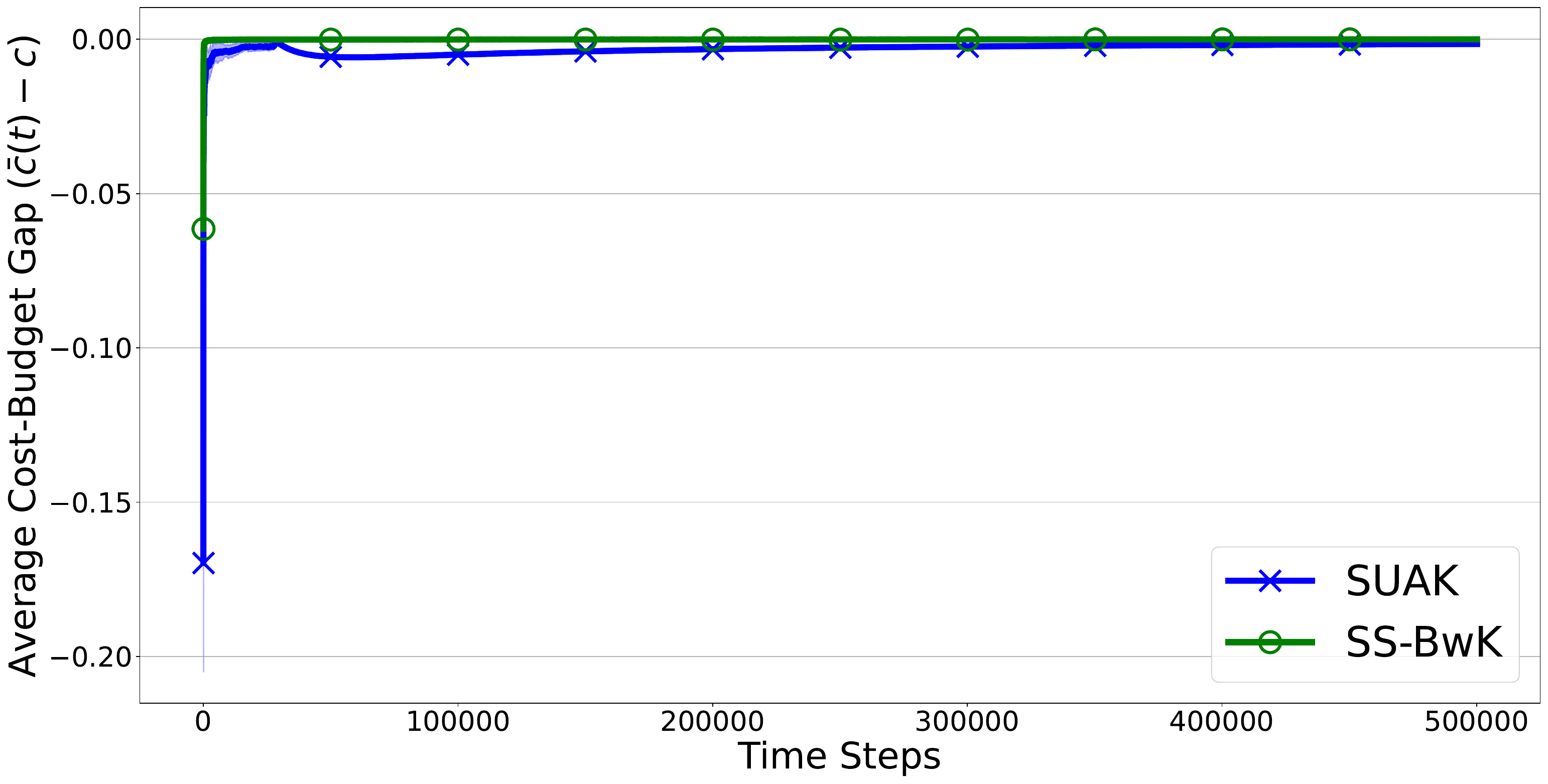}

\caption{Plots of cumulative empirical regret (top left), number of skips (top right), and average empirical cost deviation $\bar c(t)-c$ (bottom).}
 
\label{fig:sim_2}

\end{figure*}

To show that this naive approach might suffer a large regret due to large number of skips, we run simulations on the following problem instance with $|\mathcal A|=4$ arms (three non-null arms and the null arm) where $\mu = [0.45, 0.7, 0.8, 0]$; and $\rho = [0.3, 0.75, 0.8, 0]$. Except for the null arm, the arm reward (cost) values are independently sampled from Beta distributions with parameters $\alpha = 10 \mu, \beta = 10 (1-\mu)$ ($\alpha = 10 \rho, \beta = 10 (1-\rho)$). The average cost budget per round is $c=0.5$. We average results from $10$ simulation runs, each with $500,000$ rounds.
We compare the results with results from SUAK which we propose in \S \ref{sec:algorithm}. 
The simulation results are given in Figure \ref{fig:sim_2}. The shaded areas in the plots represent error bars with one standard deviation. 

It can be seen that the number of skips of SS-BwK exhibits sublinear growth and it is much higher than those of SUAK, which results in higher regret compared to SUAK. 
Hence, this shows that merely adding skips to a BwK algorithm and treating it as a BwAK algorithm is not a viable solution. 
In view of this, in the next section, we present SUAK, an algorithm that strategically under-utilizes the available budget to reduce the number of skips needed, and achieve smaller regret.

\subsection{The SUAK Algorithm}
\label{sec:algorithm}

We propose an algorithm called {\em Strategic Under-utilization for Anytime Knapsacks} (SUAK) that utilizes upper and lower confidence bounds of the arm rewards and costs using the UCB principle to upper bound the reward that can be obtained from a particular base. SUAK also uses skipping a round to satisfy the anytime cost constraint, and targets an average cost of $c - \log t /(\omega^2(t) \cdot t)$ to limit the number of skips, where $\omega(t)$ is defined in  \eqref{eq:omega_t} and depends on the LCB of the estimated minimum cost gap of arms at round $t$. The pseudo-code is provided in Algorithm \ref{alg:alg_SUAK_final}. 

SUAK works as follows. At each round, it uses the observations collected so far to construct confidence bounds on the mean rewards and costs of the arms. It then determines whether additional exploration is needed to identify which arms have mean cost below or above the budget threshold $c$. On rounds when every current cost-confidence interval excludes $c$, SUAK constructs an optimistic LP using upper confidence bounds on rewards and lower confidence bounds on costs, and selects an arm from an optimal LP base. Throughout this process, SUAK uses strategic under-utilization of the available budget and skips rounds when necessary to ensure that the anytime cost constraint is satisfied. In particular, when operating according to an LP base, it targets an average cost of $c-\log t/(\omega^2(t)t)$, where $\omega(t)$ is defined in \eqref{eq:omega_t}, in order to build sufficient budget slack and limit the number of global skips.

Before solving the LP, SUAK first determines whether the mean cost of each non-null arm can be confidently classified as lying strictly above or below the target cost $c$. An arm $l$ is \emph{cost-uncertain} at round $t$ if its broad cost confidence interval contains $c$, i.e.,
$\varrho_l^L(t)\le c\le \varrho_l^U(t),$
where
\begin{align}
\varrho_l^L(t)
&:= \Bar{\rho}_l(t)-7\sqrt{\frac{1.5\log t}{N_l(t)}},\\
\varrho_l^U(t)
&:= \Bar{\rho}_l(t)+7\sqrt{\frac{1.5\log t}{N_l(t)}}.
\label{def}
\end{align}
For an unobserved non-null arm, we set
$\varrho_i^L(t)=-\infty,
\ 
\varrho_i^U(t)=+\infty,$
and
$\rho_i^L(t)=\mu_i^L(t)=0,\ 
\rho_i^U(t)=\mu_i^U(t)=1,$
without evaluating any expression involving $1/N_i(t)$ when $N_i(t)=0$. To separately account for the cost incurred while resolving this cost uncertainty, SUAK maintains a dedicated classification cost tracker. Let $S_p(t)$ denote the cumulative realized cost of classification pulls through round $t$, and let $N_p(t)$ denote the number of classification rounds, including both classification pulls and the dedicated skips used to keep the average cost of classification rounds below $c$. We initialize
$S_p(0)=N_p(0)=0.$

\begin{algorithm}[!ht]
\caption{SUAK: Strategic Under-utilization for Anytime Knapsacks}
\label{alg:alg_SUAK_final}
\begin{multicols}{2}
\begin{algorithmic}[1]
\STATE \textbf{Input: } Average cost target $0<c<1$
\STATE Initialize $S_c(0)=S_p(0)=N_p(0)=0$ and $N_i(1)=0$ for $i\in[K]$
\FOR{each round $t\ge1$}
    \STATE Form round-$t$ confidence bounds from observations before $t$; keep null-arm bounds equal to zero
    \STATE $S_c(t)\gets S_c(t-1)$; $S_p(t)\gets S_p(t-1)$; $N_p(t)\gets N_p(t-1)$, $\chi_t^{\rm cls}\gets0$
    \IF{$S_p(t-1)+1>cN_p(t-1)$}
        \STATE $i(t)\gets \bot$; $N_p(t)\gets N_p(t-1)+1$ \alglinelabel{alg:SUAK_skip_p2}
    \ELSIF{$S_c(t-1)+1>ct$}
        \STATE $i(t)\gets \bot$ \alglinelabel{alg:SUAK_skip_c}
    \ELSIF{$\exists l\in[K]:\varrho_l^L(t)\le c\le\varrho_l^U(t)$} \alglinelabel{alg:e3_conf_pull_cond1}
        \STATE Choose such an $l$ with deterministic tie-breaking; $i(t)\gets l$ \alglinelabel{alg:e3_conf_pull_cond}
        \STATE $\chi_t^{\rm cls}\gets1$; $N_p(t)\gets N_p(t-1)+1$ \alglinelabel{alg:e3_conf_pull_cond2}
    \ELSE
        \STATE Evaluate $\delta_{\min}^L(t)$ and $\omega(t)$ using \eqref{eq:deltamin_t} and \eqref{eq:omega_t}
        \STATE $\mathcal S_t\gets\{\pi\in\Delta_{K+1}:\langle\pi,\boldsymbol\rho^L(t)\rangle\le c\}$
        \STATE Choose an optimal basic feasible solution $\pi(t)$ of $\max_{\pi\in\mathcal S_t}\langle\boldsymbol\mu^U(t),\pi\rangle$, with deterministic tie-breaking
        \STATE $\mathcal I_t\gets\{q:\pi_q(t)>0\}$
        \IF{$|\mathcal I_t|=1$}
            \STATE Set $i(t)$ to the unique member of $\mathcal I_t$
        \ELSE
            \STATE Write $\mathcal I_t=\{j,k\}$ with $\bar\rho_j(t)>\bar\rho_k(t)$
            \STATE $b(t)\gets ct-S_c(t-1)-\log t/\omega^2(t)$
            \IF{$b(t)>\bar\rho_j(t)$}
                \STATE $p(t)\gets1-\omega(t)$
            \ELSIF{$b(t)<\bar\rho_k(t)$}
                \STATE $p(t)\gets\omega(t)$
            \ELSE
                \STATE $p_1(t)\gets\max\{(b(t)-\bar\rho_k(t))/(\bar\rho_j(t)-\bar\rho_k(t)),\omega(t)\}$
                \STATE $p(t)\gets\min\{p_1(t),1-\omega(t)\}$
            \ENDIF
            \STATE Draw $i(t)=j$ with probability $p(t)$ and $i(t)=k$ otherwise
        \ENDIF
    \ENDIF
    \STATE Pull arm $i(t)$ and observe $r(t),c(t)$ \alglinelabel{alg:e4_base_pull}
    \STATE $S_c(t)\gets S_c(t)+c(t)$
    \STATE $S_p(t)\gets S_p(t)+\chi_t^{\rm cls}c(t)$
    \STATE Update $N_i(t+1)$ and empirical means for round $t+1$, for every $i\in[K]$
\ENDFOR
\end{algorithmic}
\end{multicols}
\end{algorithm}

 At each round, SUAK performs the following steps. SUAK first checks if the empirical average cost of classification pulls exceed $c$, i.e. if $S_p(t-1)+1>cN_p(t-1)$; and, skips that round in that case. Otherwise, it next checks the global anytime constraint and executes a global skip whenever
$S_c(t-1)+1>ct.$ 
After these checks, SUAK tests for cost uncertainty. If some non-null arm has a broad cost confidence interval containing $c$, one such arm is selected for classification (any tie-braker can be used if there are multiple such arms); otherwise, SUAK proceeds to the LP step. Note that since these intervals are recomputed every round, the classification condition may become active again later. 

Whenever SUAK reaches the LP step, it computes a lower confidence bound on the minimum cost gap and the corresponding under-budgeting parameter:
\begin{align}
\delta_{\min}^L(t)
&:=
\min_{i\in[K]}
\left(
|\bar{\rho}_i(t)-c|
-\sqrt{\frac{1.5\log t}{N_i(t)}}
\right),
\label{eq:deltamin_t}\\
\omega(t)
&:=
\frac{\delta_{\min}^L(t)}
{2+\delta_{\min}^L(t)-c}.
\label{eq:omega_t}
\end{align}
It then forms the optimistic feasible set
\[
\mathcal S_t
=
\left\{
\boldsymbol{\pi}\in\Delta_{K+1}:
\langle\boldsymbol{\pi},\boldsymbol{\rho}^L(t)\rangle\le c
\right\},
\]
where
\begin{align}
\mu_i^U(t)
&:=
\operatorname{proj}_{[0,1]}
\left(
\Bar{\mu}_i(t)+\epsilon_i(t)
\right),\\
\rho_i^L(t)
&:=
\operatorname{proj}_{[0,1]}
\left(
\Bar{\rho}_i(t)-\epsilon_i(t)
\right),
\end{align}
and
\[
\epsilon_i(t)
=
\sqrt{\frac{3\log t}{N_i(t)}}.
\]
The corresponding upper cost bound $\rho_i^U(t)$ and lower reward bound $\mu_i^L(t)$ are defined analogously.

SUAK then solves the optimistic LP
\[
\max_{\pi\in\mathcal S_t}
\langle\boldsymbol{\mu}^U(t),\pi\rangle.
\]
If several optimal solutions exist, SUAK selects an optimal basic feasible solution using deterministic tie-breaking. Because the LP has a single cost constraint in addition to the simplex constraint, such a solution has support size at most two. Let
$\mathcal I_t
=
\{q:\pi_q(t)>0\}$
denote the support of the selected solution. If $|\mathcal I_t|=1$, SUAK pulls the unique arm in the support. If $|\mathcal I_t|=2$, write
\[
\mathcal I_t=\{j(t),k(t)\},
\qquad
\Bar{\rho}_{j(t)}(t)>\Bar{\rho}_{k(t)}(t),
\]
so that $j(t)$ is the empirically higher-cost arm.

For a selected two-arm base, SUAK defines the effective round-$t$ budget as
\[
b(t)
=
ct-S_c(t-1)-\frac{\log t}{\omega^2(t)}.
\]
where the term $\log t/\omega^2(t)$ is the under-budgeting amount. If
$b(t)>\Bar{\rho}_{j(t)}(t),$
SUAK chooses the higher-cost arm with probability $1-\omega(t)$. If
$b(t)<\Bar{\rho}_{k(t)}(t),$
it chooses the higher-cost arm with probability $\omega(t)$. Otherwise, it uses
\[
p(t)
=
\frac{
b(t)-\Bar{\rho}_{k(t)}(t)
}{
\Bar{\rho}_{j(t)}(t)-\Bar{\rho}_{k(t)}(t)
},
\]
clipped to the interval
$[\omega(t),1-\omega(t)].$
The higher-cost arm is then selected with probability $p(t)$ and the lower-cost arm otherwise. This clipping guarantees that, on the relevant good events, each arm in a selected two-arm base is sampled with probability $\omega(t)$. A nonzero lower bound for $\omega(t)$ is established in Lemma~\ref{cor:delta_omega_estimate}.

Note that SUAK is non-stationary as it is adaptive to the available budget at that round. This design is essential as it was shown in \citet[Lemma 2]{flajolet2015logarithmic_knap_2} that a non-adaptive design suffers $\Omega(\sqrt{T})$ regret even if the optimal solution $\boldsymbol{\pi}^* $ is known unless all arms consume the same deterministic cost every round. The main intuition behind this result is that the fluctuation of the available budget around its mean at round $t$ can be as high as $\Omega(1/\sqrt{t})$.

\section{Analysis of SUAK}
\label{sec:analysis}

In this section we provide an instance-dependent regret analysis of \textsc{SUAK}. The main difficulty compared with a standard UCB proof is that a selected base is only partially observed: in a two-arm base, only one of the two arms is pulled in a round. In addition, the algorithm incurs regret both from explicitly classifying arms relative to the cost threshold $c$ and from deliberately under-utilizing the budget. The analysis below separates these effects and shows that each contributes at most logarithmic regret. 

Consistently with \eqref{eq:canonical_vertex_set}, each base in the regret analysis is a basic feasible solution of the LP. Thus a two-arm base $\mathcal I=(h,\ell)$ will have $\rho_h>c>\rho_\ell$ (where $\ell$ may be the null arm), and its weight vector will be
\begin{align}
 \pi_h^{\mathcal I}
 :=\frac{c-\rho_\ell}{\rho_h-\rho_\ell},
 \qquad
 \pi_\ell^{\mathcal I}
 :=\frac{\rho_h-c}{\rho_h-\rho_\ell}.
 \label{eq:canonical_base_mix}
\end{align}
For a one-arm base $\mathcal I=\{i\}$ with $\rho_i\le c$, $\pi_i^{\mathcal I}=1$ will hold.  We define
\begin{align}
 r_{\mathcal I}:=\boldsymbol\mu^T\boldsymbol\pi^{\mathcal I}.
 \label{eq:base_value}
\end{align}
Then, the optimal reward will be
$r^*=\max_{\mathcal I\in\mathds V}r_{\mathcal I}$.  Note that for a two-arm base, \(r_{\mathcal I}\) denotes the reward of the unique mixture on the cost-\(c\) boundary, rather than the maximum reward over all feasible policies whose support is contained in \(\mathcal I\). This distinction is important for the regret decomposition developed below.

\subsection{Gap-Dependent Regret Bound}
\label{subsec:gap_dependent}

\begin{definition}[Reward gaps]
For an arm $i\in[K]$, let $\Delta_i:=\mu_{\bestarm}-\mu_i$. For a base $\mathcal I\in\mathds V$, we define
\begin{align}
    \Delta_{\mathcal I}:=r^*-r_{\mathcal I}.
\end{align}
For each arm $i$, we define the smallest positive base gap among bases containing $i$ as
\begin{align}
\Delta_{\min,i}:=
\min\{\Delta_{\mathcal I}:\mathcal I\in\mathds V,\ i\in\mathcal I,\ \Delta_{\mathcal I}>0\},
\label{eq:delta_min_i_def}
\end{align}
with the convention $\Delta_{\min,i}=\infty$ if no suboptimal base contains $i$.
\end{definition}

\begin{definition}[Cost gaps]
For each arm $i\in[K]$, let
\begin{align}
\delta_i:=|\rho_i-c|,\qquad \delta_{\min}:=\min_{i\in[K]}\delta_i.
\label{eq:cost_gaps}
\end{align}
\end{definition}
For notational convenience in bases containing the null arm, we set
\begin{align}
 \rho_{\bot}=\mu_{\bot}=0,\qquad
 \delta_{\bot}:=c,\qquad
 \epsilon_{\bot}(t):=0,\qquad N_{\bot}(t):=\infty.
\end{align}

\begin{definition}[Good event]
For each round \(t\), define the confidence radii
\begin{align}
 a_i(t):=\sqrt{\frac{1.5\log t}{N_i(t)}},
 \qquad
 \epsilon_i(t):=\sqrt{\frac{3\log t}{N_i(t)}}.
\end{align}
Using these, we define the good event \(\mathcal G_t\) as
\begin{align}
\mathcal G_t
:=
\left\{
\forall i\in[K]\text{ with }N_i(t)>0:
\begin{array}{l}
|\Bar\rho_i(t)-\rho_i|\le a_i(t),\\[1mm]
|\Bar\mu_i(t)-\mu_i|\le \epsilon_i(t)
\end{array}
\right\}.
\label{eq:good_event}
\end{align}
We also define the number of rounds up to \(T\) on which this event fails as
\begin{align}
 F_T:=\sum_{t=1}^T\ind{\mathcal G_t^c}.
 \label{eq:failure_count_def}
\end{align}
A concentration bound on \(F_T\) is established immediately before the proof of Theorem~\ref{thm:gap_dependent_suak}.
\end{definition}

\begin{assumption}
\label{ass:positive_cost_gap}
We assume $\delta_{\min} > 0$.
\end{assumption}

Assumption~\ref{ass:positive_cost_gap} is necessary since if $\rho_i=c$ for some arm, no finite number of samples can determine on which side of $c$ that arm lies with a fixed positive margin, and the classification and under-budgeting terms below become singular.

\begin{lemma}[Uniform exploration probability]
\label{lem:omega_bounds}\label{cor:delta_omega_estimate}
Under Assumption~\ref{ass:positive_cost_gap}, on $\mathcal G_t$, whenever the LP step of \textsc{SUAK} is reached,
\begin{align}
    \omega_{\min}:=\frac{2}{9}\delta_{\min}
    \le \omega(t)
    \le \frac{\delta_{\min}}{2+\delta_{\min}-c}
    =:\omega_{\max}
    \le \delta_{\min}.
\label{eq:omega_bounds}
\end{align}
Hence, an arm belonging to a selected two-arm base is pulled with probability at least $\omega_{\min}$.
\end{lemma}
Proof for this result is provided in Appendix~\ref{app:proof_lem_omega_bounds}.

\begin{lemma}[Dual-price regret decomposition]
\label{lem:dual_decomposition}
Let $(\nu^*,\lambda^*)$ be an optimal solution of the dual LP
\begin{align}
 r^*=\min_{\nu\in\mathbb R,\,\lambda\ge0}\;&\nu+c\lambda \\
 \text{s.t. }&\nu+\lambda\rho_i\ge\mu_i,
 \qquad i\in\mathcal A.
 \label{eq:dual_lp}
\end{align}
Define the reduced-cost gap
\begin{align}
 g_i:=\nu^*+\lambda^*\rho_i-\mu_i\ge0.
 \label{eq:reduced_cost_gap}
\end{align}
Then $0\le \nu^*\le 1$ and $\lambda^*\le1/c$. Moreover, for every two-arm base $\mathcal I$,
\begin{align}
 \Delta_{\mathcal I}
 =\sum_{q\in\mathcal I}\pi_q^{\mathcal I}g_q,
 \label{eq:base_gap_reduced_cost}
\end{align}
while for a one-arm base $\mathcal I=\{i\}$, $g_i\le\Delta_{\mathcal I}$. Define the structural charge as
\begin{align}
 G_{\mathcal I}:=
 \begin{cases}
\max_{q\in\mathcal I}g_q,& |\mathcal I|=2,\\
  \Delta_{\mathcal I},& |\mathcal I|=1.
 \end{cases}
\label{eq:structural_charge}
\end{align}
Then $G_{\mathcal I}\ge\Delta_{\mathcal I}$, and $G_{\mathcal I}>0$ if and only if $\Delta_{\mathcal I}>0$.
Consequently, on an ordinary LP round $t$ for which the selected support is a canonical true base (in particular, on $\mathcal G_t$),
\begin{align}
 r^*-\E[r(t)\mid\mathcal F_{t-1}]
 \le
 G_{\mathcal I_t}
 +\lambda^*\!
 \left(c-\E[c(t)\mid\mathcal F_{t-1}]\right).
 \label{eq:dual_round_decomposition}
\end{align}
For an optimal selected base the first term on the right is zero. Further, if $|\mathcal I_t|=1$, the expected regret is $G_{\mathcal I_t}$.
\end{lemma}
The proof is provided in Appendix~\ref{app:proof_lem_dual_decomposition}.

\begin{lemma}[Cost-classification complexity]
\label{lem:cost_classification}
Let $N_i^{{\rm cls},G}(T)$ be the number of pulls of arm $i$ caused by the condition
$\varrho_i^L(t)\le c\le\varrho_i^U(t)$ on rounds for which $\mathcal G_t$ holds. Then
\begin{align}
    N_i^{{\rm cls},G}(T)\le \frac{96\log T}{\delta_i^2}+1.
\label{eq:classification_pulls}
\end{align}
Let $M_{\rm cls}(T)$ be the total number of classification pulls up to $T$, including pulls on confidence-failure rounds, and let $S_{\rm cls}(T)$ be the number of rounds skipped due to the classification cost tracker. Pathwise,
\begin{align}
 S_{\rm cls}(T)
 \le \frac{M_{\rm cls}(T)}{c}+\frac1c+1.
\label{eq:classification_skips}
\end{align}
Consequently,
\begin{align}
 \E[M_{\rm cls}(T)]
 \le 96\sum_{i=1}^K\frac{\log T}{\delta_i^2}+K+\E[F_T],
\label{eq:classification_total_pulls}
\end{align}
where $F_T:=\sum_{t=1}^T\ind{\mathcal G_t^c}$.
\end{lemma}
The proof is provided in Appendix~\ref{app:proof_lem_cost_classification}.

The next lemma considers the number of suboptimal base selections from the LP stage. Note that cost classification pulls play a crucial here, by the time a base is selected, the cost confidence interval of each arm in that base is already sufficiently narrow relative to the arm’s distance from \(c\). As a result, the optimistic LP cannot perturb the corresponding mixing weights by an arbitrarily large amount.

\begin{lemma}[Reduced-cost charging for suboptimal bases]
\label{lem:suboptimal_base}
Let $\mathcal I$ be a suboptimal canonical base selected on the good event $\mathcal G_t$. Then at least one non-null arm $q\in\mathcal I\cap[K]$ satisfies
\begin{align}
 N_q(t)
 \le
 \frac{432\log t}{\delta_q^2G_{\mathcal I}^2}.
\label{eq:base_sample_threshold}
\end{align}
Associate each good selection of a suboptimal base with one of its under-sampled non-null arms satisfying \eqref{eq:base_sample_threshold}, using deterministic tie-breaking. For each non-null arm $i$, define
\begin{align}
 G_{\min,i}:=\min\{G_{\mathcal I}:\mathcal I\in\mathds V,\ i\in\mathcal I,\ G_{\mathcal I}>0\},
 \label{eq:G_min_i_def}
\end{align}
with an empty minimum equal to infinity. Then $G_{\min,i}\ge\Delta_{\min,i}$, and the expected sum $R_{{\rm str},i}(T)$ of charges $G_{\mathcal I_t}$ assigned to $i$ satisfies
\begin{align}
 R_{{\rm str},i}(T)
 \le
 \frac{3888\log T}{\delta_{\min}\delta_i^2G_{\min,i}}
 \le
 \frac{3888\log T}{\delta_{\min}\delta_i^2\Delta_{\min,i}}.
\label{eq:arm_charged_base_regret}
\end{align}
These charges jointly upper-bound the expected reduced-cost losses on good two-arm base LP rounds and the expected regrets on good single-arm base LP rounds.
\end{lemma}
The proof is provided in Appendix~\ref{app:proof_lem_suboptimal_base}.

\begin{lemma}[Joint two-arm and global-skip cost tracking]
\label{lem:skip_underbudget}
Let $\mathcal T_2^{\rm G}(T)$ be the set of LP rounds up to $T$ in which a two-arm base is selected and $\mathcal G_t$ holds. Let $J_t^{\rm gskip}$ denote the indicator that \textsc{SUAK} actually executes the global anytime-skip branch at round $t$ (line~\ref{alg:SUAK_skip_c}).Since the check of classification cost tracker is evaluated first, it holds that
\begin{align}
 J_t^{\rm gskip}
 :=\ind{S_p(t-1)+1\le cN_p(t-1)}
   \ind{S_c(t-1)+1>ct}.
\label{eq:gskip_branch_indicator}
\end{align}
Define
\begin{align}
 N_{\rm gskip}^{\rm G}(t):=
 \sum_{s=1}^t\ind{\mathcal G_s}J_s^{\rm gskip},
\end{align}
and let $F_t:=\sum_{s=1}^t\ind{\mathcal G_s^c}$. Define
\begin{align}
 Q_t:=
 \sum_{s\le t:\,s\in\mathcal T_2^{\rm G}(T)}(c-c(s))
 +cN_{\rm gskip}^{\rm G}(t),
 \qquad
 \widetilde Q_t:=Q_t-F_t.
\label{eq:joint_Q_def}
\end{align}
Then, for $T\ge3$,
\begin{align}
 \E[Q_T]
 \le \frac{41\log T}{\delta_{\min}^2}+\E[F_T].
\label{eq:joint_cost_tracking}
\end{align}
Equivalently,
\begin{align}
 \E\!\left[
 \sum_{t\in\mathcal T_2^{\rm G}(T)}(c-\rho_{i(t)})
 +cN_{\rm gskip}^{\rm G}(T)
 \right]
 \le \frac{41\log T}{\delta_{\min}^2}+\E[F_T].
\label{eq:joint_cost_tracking_mean}
\end{align}
\end{lemma}
The proof is provided in Appendix~\ref{app:proof_lem_skip_underbudget}.

\begin{lemma}[Protective-drift bound for global skips]
\label{lem:global_skip_count}
Fix an optimal primal-dual pair $(\pi^*,\nu^*,\lambda^*)$. Define
\begin{align}
 \eta:=\omega_{\max}
 =\frac{\delta_{\min}}{2+\delta_{\min}-c}>0.
\end{align}
Since $0<c<1$ and $0<\delta_{\min}<1$, we have $\eta\ge\delta_{\min}/3$ and $\eta<1/2$.

If $\nu^*>0$, then
\begin{align}
 \E[N_{\rm gskip}^{\rm G}(T)]
 \le
 \frac{27}{c\eta}
 \left(1+\E[M_{\rm cls}(T)]+\E[F_T]\right).
\label{eq:global_skip_count}
\end{align}
Consequently, when $\nu^*>0$,
\begin{align}
 \E[N_{\rm gskip}^{\rm G}(T)]
 =O\!\left(\frac{K\log T}{c\delta_{\min}^3}\right)+O(1)
\label{eq:global_skip_count_simplified}
\end{align}
for every fixed instance. If $\nu^*=0$, no bound on $N_{\rm gskip}^{\rm G}(T)$ is needed in the regret proof, because the residual reduced-cost term $\nu^*N_{\rm gskip}^{\rm G}(T)$ vanishes identically.
\end{lemma}
The proof is provided in Appendix~\ref{app:proof_lem_global_skip_count}.

\begin{theorem}[Gap-dependent regret of \textsc{SUAK}]
\label{thm:gap_dependent_suak}
Under Assumption~\ref{ass:positive_cost_gap}, for any $T\ge3$, the expected regret of \textsc{SUAK} satisfies
\begin{align}
R_T
\le {}&
\sum_{i=1}^{K}
\frac{3888\log T}{\delta_{\min}\delta_i^2\Delta_{\min,i}}
+\frac{196Kr^*\log T}{c\delta_{\min}^2}
\nonumber\\
&\quad
+\frac{41\lambda^*\log T}{\delta_{\min}^2}
+\frac{27\nu^*}{c\eta}
\left(
 96K\frac{\log T}{\delta_{\min}^2}+6K+1
\right)
\nonumber\\
&\quad
+\left(1+\lambda^*+r^*\left(1+\frac1c\right)\right)\frac{5K}{2},
\label{eq:main_gap_regret}
\end{align}
where $(\nu^*,\lambda^*)$ is an optimal dual solution in \eqref{eq:dual_lp} and
$\eta=\delta_{\min}/(2+\delta_{\min}-c)$. Since $\lambda^*\le1/c$, $\nu^*\leq 1$, and $\eta\ge\delta_{\min}/3$, every fixed instance satisfies
\begin{align}
 R_T=O(K\log T)+O(1).
\end{align}
If $\nu^*=0$ (in particular, whenever the optimal LP uses the null arm), the global-skip-count term vanishes and the budget-dependent logarithmic term retains the $O(\lambda^*\log T/\delta_{\min}^2)$ dependence from Lemma~\ref{lem:skip_underbudget}.
\end{theorem}

\begin{proof}
We first control the confidence-failure rounds. Since $a_i(t)\le\epsilon_i(t)$, the cost condition in \eqref{eq:good_event} also implies that the true cost mean lies inside the wider cost confidence intervals used later in the proof, including those of radii $\epsilon_i(t)$ and $7a_i(t)$. For each arm, consider its observations in pull order. Due to the i.i.d assumption in \eqref{eq:joint_freshness}, this pull-order observation sequence is also i.i.d. with the marginal reward and cost distributions of the arm. A union bound over deterministic sample counts $1\le n\le t-1$ therefore avoids conditioning on a random sample count. Hoeffding's inequality gives, for $t\ge2$,
\begin{align}
 \Pr(\mathcal G_t^c)
 &\le 2K(t-1)t^{-3}+2K(t-1)t^{-6}\\
 &\le 2K(t^{-2}+t^{-5}).
 \label{eq:confidence_failure_probability}
\end{align}
Using $\int_1^\infty x^{-2}\,dx=1$ and $\int_1^\infty x^{-5}\,dx=1/4$,
\begin{align}
 \E[F_T]
 &\le 2K\sum_{t=2}^{\infty}(t^{-2}+t^{-5})
 \le \frac{5}{2}K.
 \label{eq:confidence_failure_total}
\end{align}
Thus, confidence failures contribute only a horizon-independent term linear in $K$, and we prove the gap-dependent bounds on the events $\mathcal G_t$. 
We account separately for good LP rounds, classification pulls and their dedicated skips, good global anytime skips, and confidence-failure rounds. A confidence-failure round may overlap with one of the action categories; any resulting double counting only enlarges the upper bound.

We first consider the good LP rounds. For a two-arm base, Lemma~\ref{lem:dual_decomposition} gives
\begin{align}
 r^*-\E[r(t)\mid\mathcal F_{t-1}]
 \le
 G_{\mathcal I_t}
 +\lambda^*\left(c-\E[c(t)\mid\mathcal F_{t-1}]\right).
\label{eq:theorem_two_arm_decomp}
\end{align}
For a single-arm base, the expected regret is exactly $G_{\mathcal I_t}=\Delta_{\mathcal I_t}$. Let $\mathcal T_{\rm LP}^{\rm G}(T)$ denote all good LP rounds up to $T$. Applying Lemma~\ref{lem:suboptimal_base} jointly to single-arm and two-arm base selections gives
\begin{align}
 \E\!\left[\sum_{t\in\mathcal T_{\rm LP}^{\rm G}(T)}
 G_{\mathcal I_t}\right]
 \le
 \sum_{i=1}^{K}
 \frac{3888\log T}{\delta_{\min}\delta_i^2\Delta_{\min,i}}.
\label{eq:dual_structural_total}
\end{align}
Both types of LP rounds use the same arm observation counts in the charging argument, so this structural bound is used only once.
For a good global skip the chosen action is the null arm. Since $g_{\bot}=\nu^*$, the dual identity gives
\begin{align}
 r^*=\nu^*+\lambda^*c.
\label{eq:skip_dual_identity}
\end{align}
Therefore the $\lambda^*$ cost-deficit terms from good two-arm LP rounds and the $\lambda^*c$ portions of good global-skip regret combine exactly. Lemma~\ref{lem:skip_underbudget} bounds their sum by
\begin{align}
 \frac{41\lambda^*\log T}{\delta_{\min}^2}
 +\lambda^*\E[F_T].
\label{eq:joint_lambda_term}
\end{align}
The remaining reduced-cost loss from good global skips is
$\nu^*N_{\rm gskip}^{\rm G}(T)$. If $\nu^*=0$, this term is identically zero. If $\nu^*>0$, choose any optimal primal solution $\pi^*$; together with the displayed optimal dual solution this forms an optimal primal-dual pair, so Lemma~\ref{lem:global_skip_count}, together with Lemma~\ref{lem:cost_classification}, gives
\begin{align}
 \nu^*\E[N_{\rm gskip}^{\rm G}(T)]
 \le
 \frac{27\nu^*}{c\eta}
 \left(
 1+96K\frac{\log T}{\delta_{\min}^2}+K+2\E[F_T]
 \right).
\label{eq:nu_skip_term}
\end{align}
Here we used $\sum_i\delta_i^{-2}\le K\delta_{\min}^{-2}$. Thus \eqref{eq:nu_skip_term} holds in both cases, with both sides equal to zero when $\nu^*=0$.

For the classification process, each classification pull and each dedicated classification skip has regret at most $r^*\le1$. Lemma~\ref{lem:cost_classification} therefore yields
\begin{align}
 R_{\rm cls}(T)
 &\le r^*\E[M_{\rm cls}(T)+S_{\rm cls}(T)]\\
 &\le
 r^*\left(1+\frac1c\right)
 \left(96K\frac{\log T}{\delta_{\min}^2}+K+\E[F_T]\right)
 +r^*\left(1+\frac1c\right).
\label{eq:classification_regret_explicit_pre}
\end{align}
Because $0<c<1$, $T\ge3$, $0<\delta_{\min}<1$, and $K\ge1$,
\begin{align}
96\left(1+\frac1c\right)
K\frac{\log T}{\delta_{\min}^2}
&\le
192\frac{K\log T}{c\delta_{\min}^2},\\
\left(1+\frac1c\right)(K+1)
&\le \frac{4K}{c}
\le \frac{4}{\log3}\frac{K\log T}{c\delta_{\min}^2}.
\end{align}
Since $192+4/\log3<196$, we obtain the fully explicit bound
\begin{align}
 R_{\rm cls}(T)
 \le
 196\frac{Kr^*\log T}{c\delta_{\min}^2}
 +r^*\left(1+\frac1c\right)\E[F_T].
\label{eq:classification_regret_explicit}
\end{align}

Finally, \eqref{eq:confidence_failure_total} gives $\E[F_T]\le(5/2)K$. The direct regret on confidence-failure rounds is at most $\E[F_T]$, the joint tracking term contributes $\lambda^*\E[F_T]$, and classification cost tracking contributes $r^*(1+1/c)\E[F_T]$. This concludes the proof of the theorem.
\end{proof}

\begin{remark}[Shadow prices and reduced-cost charging]
A unit of unused cost is not generally worth $r^*/c$ units of reward; the appropriate global conversion rate is the optimal dual multiplier $\lambda^*$. The remaining loss on a selected two-arm base is bounded directly by $G_{\mathcal I}=\max_{q\in\mathcal I}g_q$, regardless of SUAK's adaptive within-base randomization. Retaining the canonical mixture weights in the optimism calculation gives $G_{\mathcal I}\le12\max_{q\in\mathcal I}\epsilon_q(t)/\delta_q$. Charging these reduced-cost gaps therefore preserves the $1/\Delta_{\min,i}$ reward-gap dependence without an additional inverse canonical-weight factor. This argument does not require SUAK to track each base's canonical occupation measure. Together with the linear protective drift $\eta\ge\delta_{\min}/3$, it yields the cubic cost-gap dependence below.
\end{remark}

\begin{corollary}[Simplified instance-dependent regret bound]\label{cor:simplified_gap_regret}
Let $\Delta_{\min}:=\min_{i:\Delta_{\min,i}<\infty}\Delta_{\min,i}$, with $\Delta_{\min}=\infty$ if no suboptimal base exists. Then Theorem~\ref{thm:gap_dependent_suak}, $r^*\le1$, $\lambda^*\le1/c$, and $\eta\ge\delta_{\min}/3$ imply
\begin{align}
R_T
=O\!\left(
\frac{K\log T}{\delta_{\min}^3\Delta_{\min}}
+\frac{\nu^*K\log T}{c\delta_{\min}^3}
+\frac{K\log T}{c\delta_{\min}^2}
\right)
+O\!\left(
K\left(
1+\frac1c
+\frac{\nu^*}{c\delta_{\min}}
\right)
\right).
\label{eq:simplified_gap_regret}
\end{align}
For fixed $c$, $\delta_{\min}$, and reward gaps this is $O(K\log T)$. The explicit cost-gap dependence of the logarithmic terms is at most cubic in $1/\delta_{\min}$, while the structural term retains the $1/\Delta_{\min}$ reward-gap dependence. Note that $0\leq \nu^* \leq 1$, and if $\nu^*=0$, the logarithmic global-skip term disappears. 
\end{corollary}

\subsection{Instance-Dependent Lower Bound}
\label{sec:lower_bound}

We next apply the classical Lai--Robbins instance-dependent lower-bound argument \citep{lai1985asymptotically} to an embedded stochastic multi-armed-bandit subclass of BwAK. The subclass used below automatically satisfies the anytime constraint, so the reduction transfers the classical logarithmic lower bound directly. Consequently, the $K\log T$ scaling of Theorem~\ref{thm:gap_dependent_suak} cannot be improved uniformly on this subclass for fixed reward gap. In particular, the argument does not characterize the information cost of learning consumptions, identifying an optimal binding-budget mixture, or avoiding budget exhaustion, and it does not provide a matching lower bound for the inverse cost-gap factors in Theorem~\ref{thm:gap_dependent_suak}.

Let $\Theta_+$ denote the class of BwAK instances satisfying Assumption~\ref{ass:positive_cost_gap}. We call a feasible policy \emph{uniformly efficient on $\Theta_+$} if, for every instance in $\Theta_+$ and every $a>0$, its regret satisfies $R_T=o(T^a)$. All uses of uniform efficiency in this subsection refer to this class. This restriction aligns the lower bound with Theorem~\ref{thm:gap_dependent_suak}; it imposes no efficiency requirement at zero-cost-gap instances. For Bernoulli parameters $p,q\in(0,1)$, define the binary relative entropy
\begin{align}
 \operatorname{kl}(p,q)
 :=p\log\frac{p}{q}+(1-p)\log\frac{1-p}{1-q}.
\label{eq:binary_kl}
\end{align}

\begin{theorem}[Classical-bandit subclass lower bound for BwAK]
\label{thm:bwak_lower_bound}
Fix any $0<\rho_0<c$, and consider the subclass of BwAK instances in which every non-null arm has deterministic cost $\rho_i(t)=\rho_0$ and Bernoulli rewards with means $\mu_i\in(0,1)$. Assume that the best arm is unique, let $\mu_*:=\max_i\mu_i$, and define $\Delta_i:=\mu_*-\mu_i$. Then every feasible policy uniformly efficient on $\Theta_+$ satisfies
\begin{align}
 \liminf_{T\to\infty}
 \frac{R_T}{\log T}
 \ge
 \sum_{i:\Delta_i>0}
 \frac{\Delta_i}{\operatorname{kl}(\mu_i,\mu_*)}.
\label{eq:bwak_lai_robbins}
\end{align}
Consequently, there exists a family of BwAK instances for which
\begin{align}
 R_T=\Omega\!\left(\frac{K\log T}{\Delta}\right),
\label{eq:bwak_symmetric_lower}
\end{align}
where $\Delta$ is the common reward gap of the suboptimal arms. In particular, for fixed $\Delta>0$, every uniformly efficient feasible policy incurs $\Omega(K\log T)$ regret on some BwAK instance.
\end{theorem}
The proof is provided in Appendix~\ref{app:proof_thm_bwak_lower_bound}.

\begin{remark}[Scope of the lower bound]
Theorem~\ref{thm:bwak_lower_bound} is the classical Lai--Robbins lower bound \citep{lai1985asymptotically} transferred to a BwAK subclass whose costs make the anytime constraint nonbinding. Accordingly, the matching statement is limited to the logarithmic order in $T$, linear dependence on $K$, and the standard $1/\Delta$ reward-gap scaling on that embedded subclass. The theorem does not establish asymptotic optimality of SUAK's coefficient, characterize regret on every fixed BwAK instance, or show that the higher-order dependence of Theorem~\ref{thm:gap_dependent_suak} on the cost gaps $\delta_i$ or $\delta_{\min}$ is information-theoretically necessary. A converse for genuinely binding-budget instances would need to capture the information required to learn consumptions and identify an optimal mixture while preserving anytime feasibility; establishing such a lower bound remains an interesting direction for future work.
\end{remark}

\section{Simulations} \label{sec:simulations}

\begin{figure*}[!t]
\centering
\includegraphics[width=0.45\textwidth]{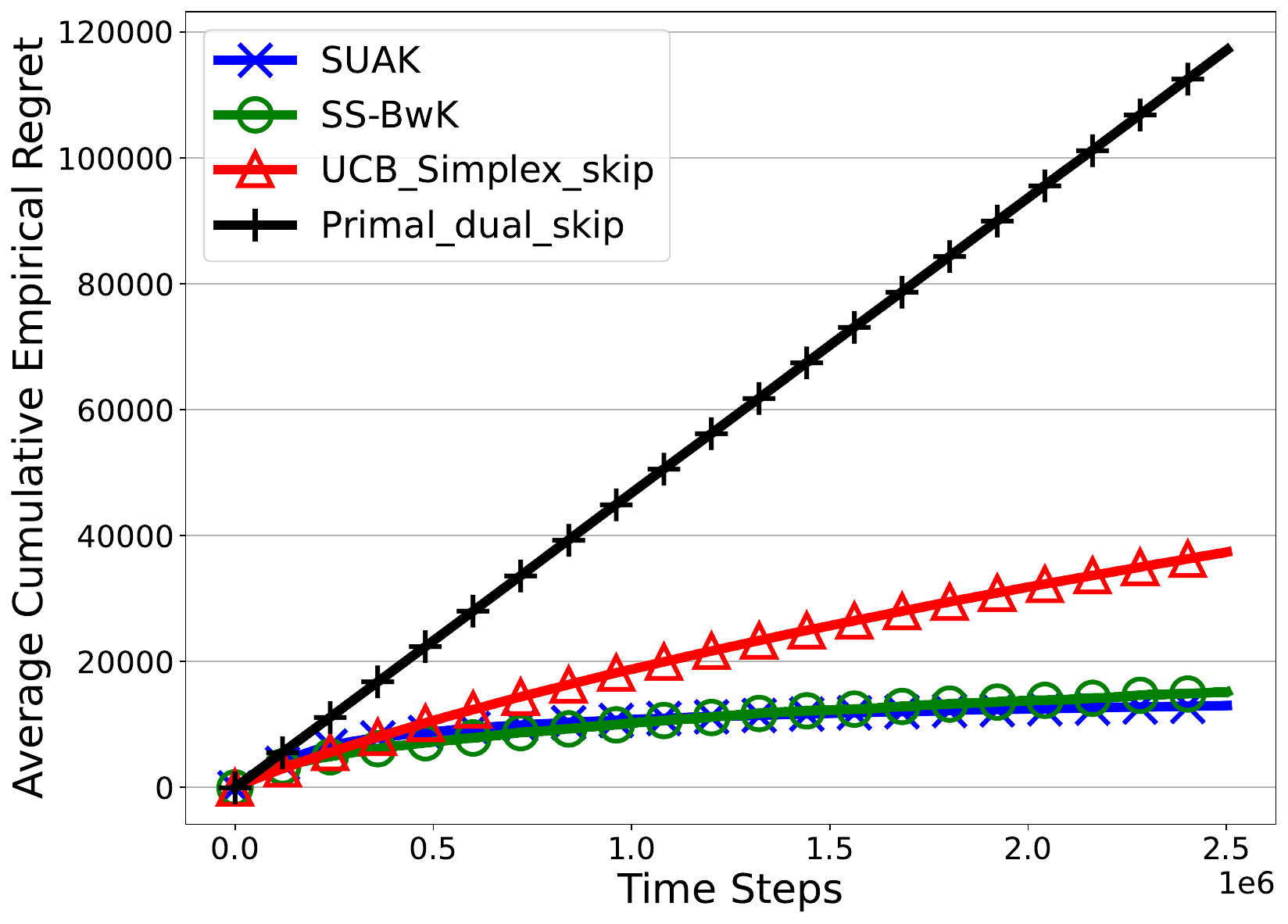} \hspace{15pt} 
\includegraphics[width=0.45\textwidth]{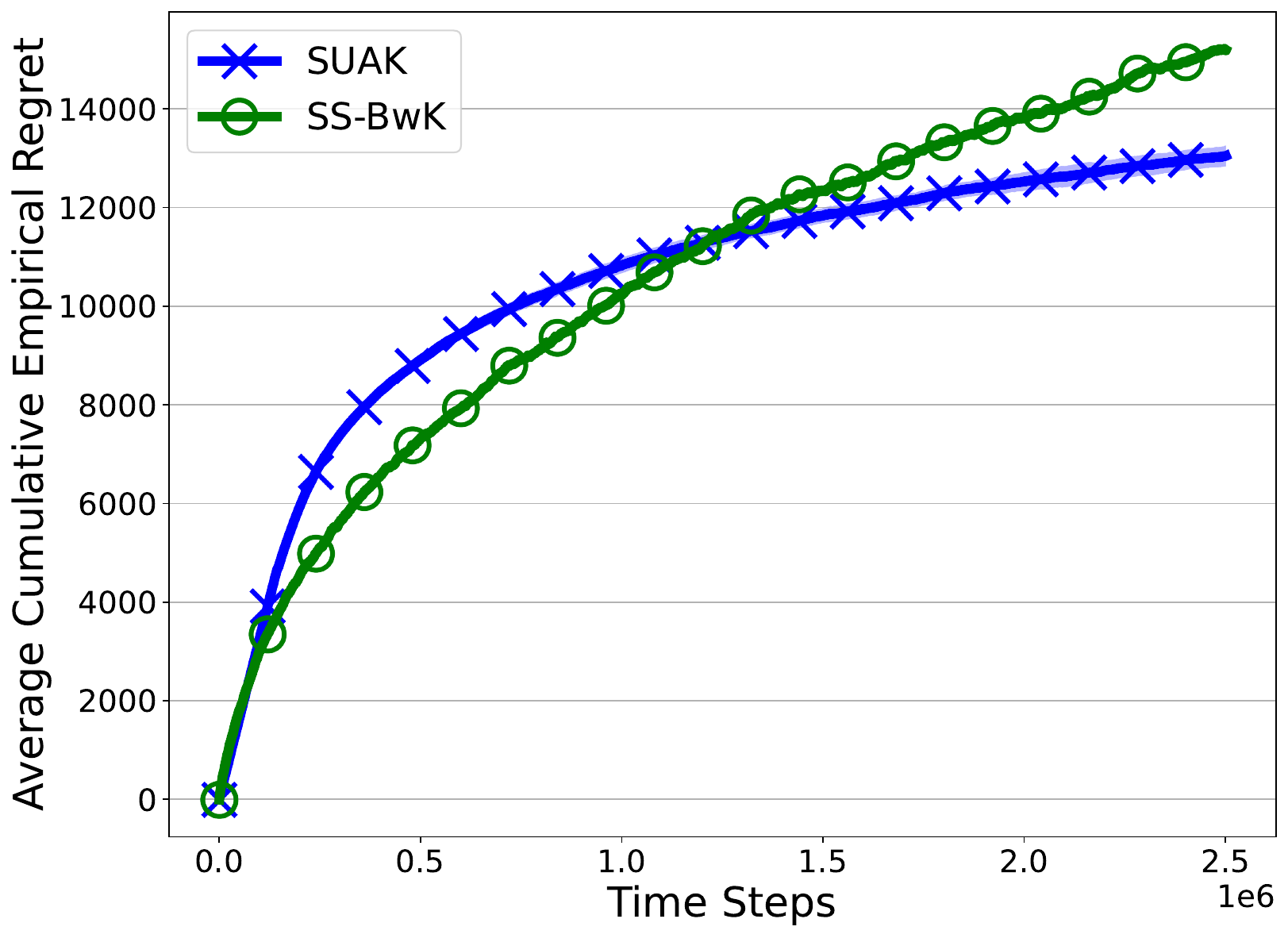} \hspace{3pt} 
\includegraphics[width=0.45\textwidth]{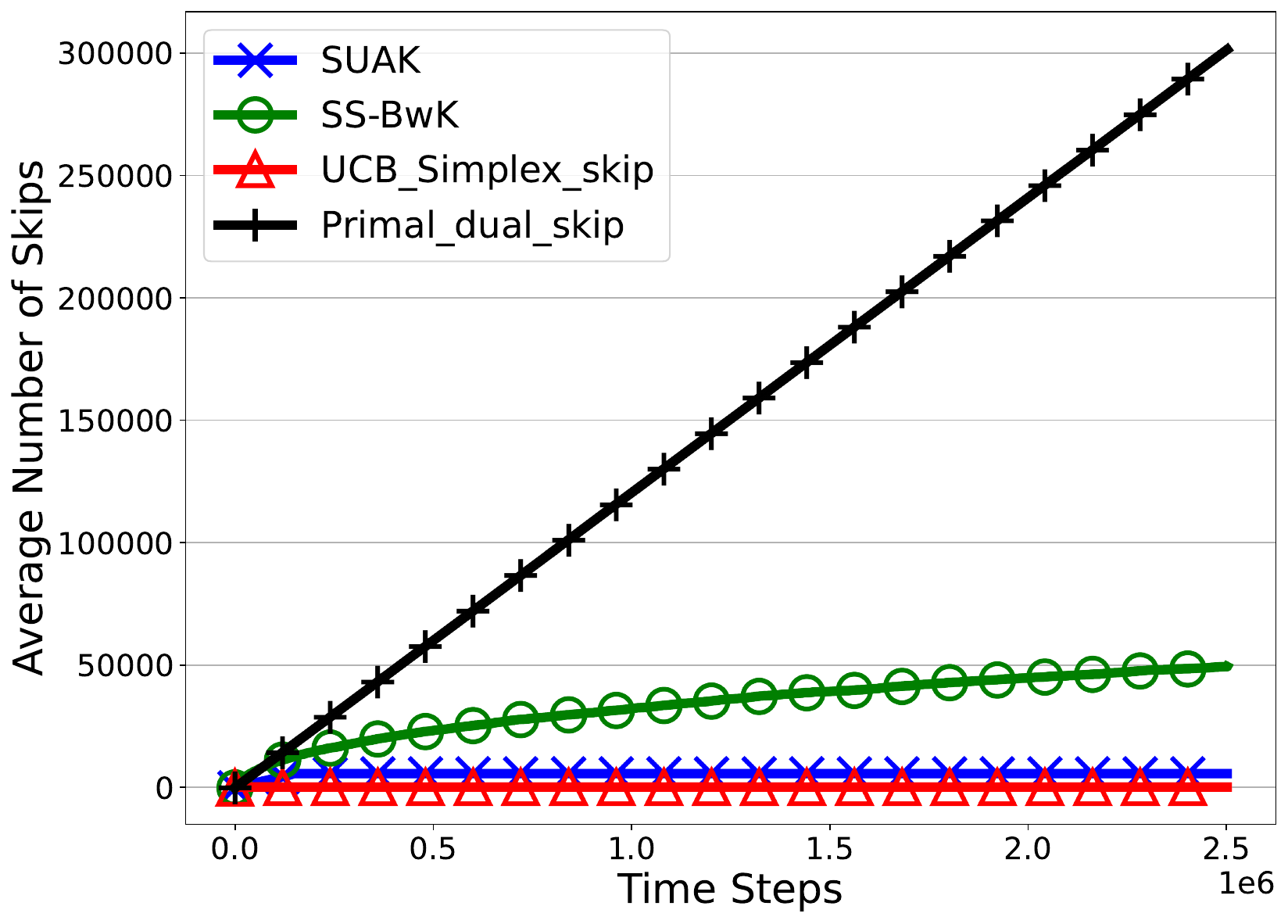} \hspace{15pt}
\includegraphics[width=0.45\textwidth]{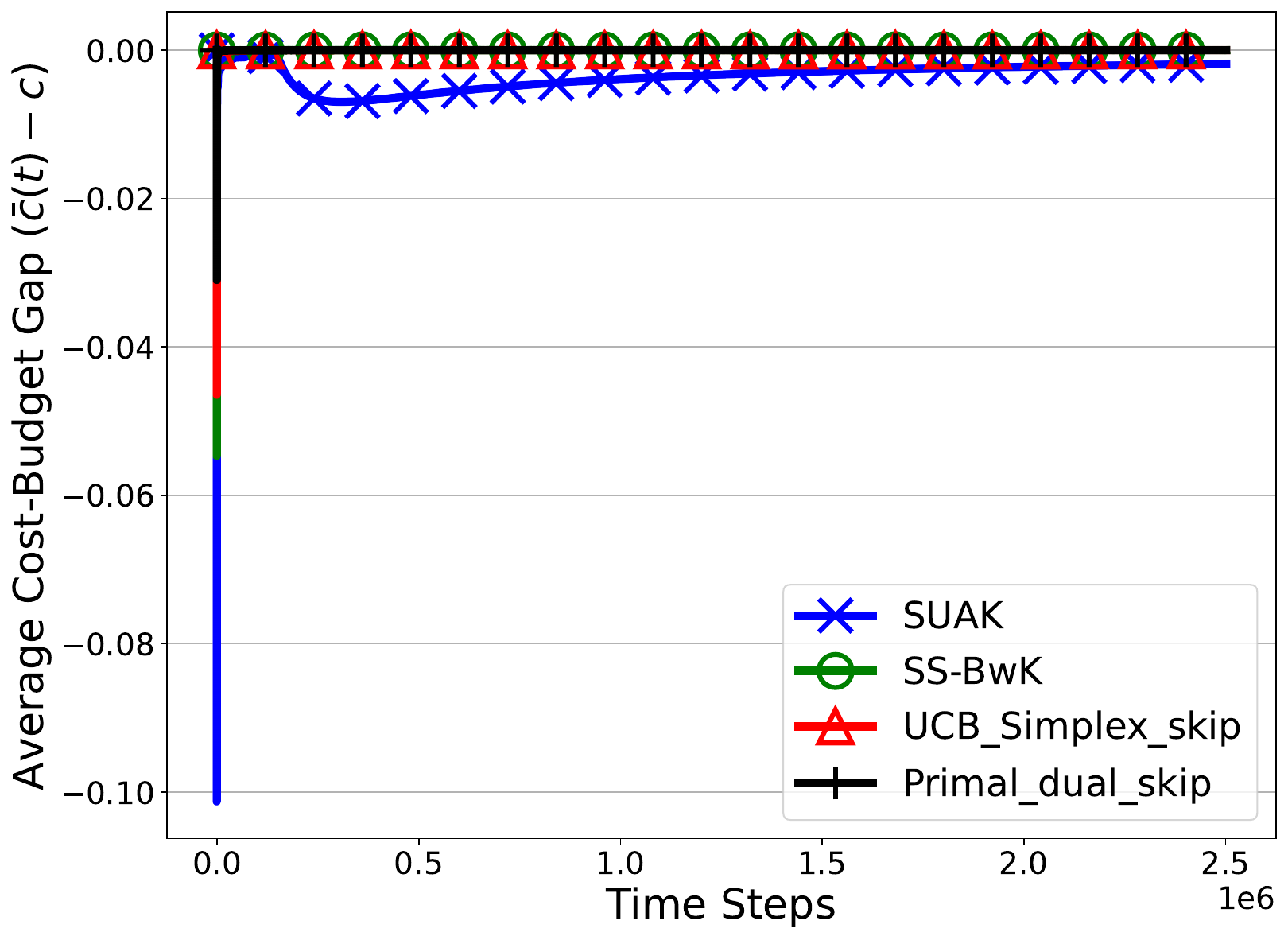}
\caption{Plots of cumulative empirical regret (top row), number of skips (bottom left), and average empirical cost deviation $\bar c(t)-c$ (bottom right). Null-arm pulls caused by dedicated or global skip branches are counted as skips; null-arm pulls selected within an LP base are not.}
\label{fig:sim_1}
\end{figure*}

We evaluate the empirical performance of SUAK through simulations. For comparison, we include the {\em Primal Dual} and {\em One Phase} algorithms of \citet{li2021symmetry_knap_1} and the {\em UCB Simplex} algorithm of \citet{flajolet2015logarithmic_knap_2}. \citet{li2021symmetry_knap_1} leave the derivation of theoretical regret guarantees for the {\em One Phase} algorithm as an open problem and instead analyze the related {\em Primal Dual} algorithm, which is expected to have weaker empirical performance. We implement their skip versions as described in \S \ref{sec:naive_algorithm}, and refer to them by appending '\_skip' to their names. We consider an instance with $K=8$ arms with means $\mu=[0.35,0.45,0.52,0.72,0.84,0.9,0.92,0.9,0]$, cost means $\rho=[0.25,0.3,0.4,0.6,0.7,0.75,0.8,0.85,0]$, and $c=0.5$. For each non-null arm, rewards and costs are sampled independently from Beta distributions with parameters $(10\mu_i,10(1-\mu_i))$ and $(10\rho_i,10(1-\rho_i))$, respectively. We run each simulation runs for $2.5$ million rounds, and report the results averaged over $10$ trials. The simulation results are given in Figure~\ref{fig:sim_1}. The shaded areas represent error bars with one standard deviation. The second plot in Figure~\ref{fig:sim_1} is a zoomed-in version of the first plot showing only SUAK and Safe-Skip BwK (SS-BwK).

The Primal Dual Skip algorithm (PDS) performs the worst in this experiment. This is expected because PDS is designed primarily for theoretical guarantees and pulls every arm the same number of times until it identifies the optimal solution. The UCB Simplex Skip algorithm (USS) also performs poorly because it assumes knowledge of a constant $\kappa$ {\em a priori} such that $|\mu_i-\mu_j|\leq \kappa |\rho_i-\rho_j|$ for any $i,j$, and the confidence intervals for the arm rewards are multiplied by a factor of $\lambda = 1 + 2 \kappa$. In our simulation setting, $\lambda=9$, which increases the number of samples needed for exploration.

It can be seen that SUAK performs better compared to PDS or USS, and exceeds the performance of SS-BwK after around round $1.4 \times 10^6$. 
 This is since the regret of SUAK concentrates mostly on the initial rounds. Pulls during initial rounds are used to reduce the uncertainty in the mean costs of arms, but are also helpful in determining the optimal base in the main algorithm, hence this reduces the regret of suboptimal base selections while searching for the optimal base. 
 After these rounds, SUAK can catch up to and eventually surpass the performance of SS-BwK due to the higher regret of SS-BwK from its high number of skips, verifying the practical utility of SUAK.

The third plot in Figure~\ref{fig:sim_1} shows that the number of skips grows sublinearly for all algorithms, with USS and SUAK requiring very few skips. This demonstrates the effectiveness of SUAK's under-budgeting mechanism. Null-arm pulls caused by the dedicated classification cost tracker skip in line~\ref{alg:SUAK_skip_p2} and the global anytime skip in line~\ref{alg:SUAK_skip_c} of Algorithm~\ref{alg:alg_SUAK_final} are counted as skips; null-arm pulls that arise because the null arm belongs to the selected LP base are not counted as a skip. The final plot in Figure~\ref{fig:sim_1} is the average cost-budget gap $\bar c(t)-c$. Under the anytime constraint this quantity must be non-positive on every round, so negative values correspond to under-utilization of the cost budget. During the early classification-heavy rounds, the dedicated accounting keeps the average cost close to $c$. As SUAK reaches the LP step more frequently, under-budgeting lowers the average empirical cost below the target and the deviation approaches zero as $\log t/(\omega^2(t)t)$ vanishes. The average empirical cost of other algorithms are very close to the
constraint, and except USS, need to skip to avoid exceeding $c$.

\section{Concluding Remarks}
\label{sec:concl}

In this paper, we study the bandits with anytime knapsacks (BwAK) problem, where we employ a stricter anytime cost constraint instead of a total cost budget. We provide SUAK, a novel algorithm that under-utilizes the available cost budget, utilizes skipping to limit the probability of violating the anytime constraint, and uses upper confidence bounds to balance exploration and exploitation. SUAK achieves an instance-dependent regret upper bound of $ O(K \log T)$. While logarithmic regret is already known for related models, including non-monotonic resource utilization \citep{kumar2022non_resource_alloc_bwk}, and evolving arm-specific budgets \citep{combes2015bandits}, direct comparison of this result is not possible due to differences in the problem setting and the variables used in the regret bounds.We also use the classical Lai--Robbins lower bound \citep{lai1985asymptotically} to show that logarithmic dependence on $T$ and linear dependence on $K$ are unavoidable for fixed reward gap under some problem instances.  
We provide simulation results to demonstrate the empirical performance of SUAK. Our work opens multiple directions for future research. One interesting future direction is to extend our results to $d$-dimensional costs. This is a challenging problem as the anytime constraint needs to be satisfied in every dimension, which can introduce additional skips. Another interesting open direction is when the rewards and costs of arms are non-stationary. In this case, satisfying the anytime cost constraint would be challenging but we conjecture it may be accomplished using a more conservative cost budget under-utilization.

\section{Acknowledgments}

This work was supported in part by the National Science Foundation through Grant \# CCF-2007834, by the Office of Naval Research through Grant \# N00014-23-1-2275, and by the CyLab Enterprise Security Initiative. The work of Cem Tekin was supported by the Turkish National Academy of Sciences Distinguished Young Scientist Award Program TUBA-GEBIP-2023 Grant and the 2024 Scientific and Technological Research Council of Türkiye Incentive Award. 

\bibliographystyle{plainnat}  
\bibliography{references}  

@article{bernasconi2024beyond,
  title={Beyond Primal-Dual Methods in Bandits with Stochastic and Adversarial Constraints},
  author={Bernasconi, Martino and Castiglioni, Matteo and Celli, Andrea and Fusco, Federico},
  journal={arXiv preprint arXiv:2405.16118},
  year={2024}
}

@article{bernasconi2024no,
  title={No-Regret is not enough! Bandits with General Constraints through Adaptive Regret Minimization},
  author={Bernasconi, Martino and Castiglioni, Matteo and Celli, Andrea},
  journal={arXiv preprint arXiv:2405.06575},
  year={2024}
}

@article{slivkins2024contextualbanditspackingcovering,
      title={Contextual Bandits with Packing and Covering Constraints: A Modular Lagrangian Approach via Regression}, 
      author={Aleksandrs Slivkins and Xingyu Zhou and Karthik Abinav Sankararaman and Dylan J. Foster},
      year={2024},
      eprint={2211.07484},
      archivePrefix={arXiv},
      primaryClass={cs.LG},
      url={https://arxiv.org/abs/2211.07484}, 
}

@inproceedings{hou2023probably_anytime_safe,
  title={Probably anytime-safe stochastic combinatorial semi-bandits},
  author={Hou, Yunlong and Tan, Vincent YF and Zhong, Zixin},
  booktitle={International Conference on Machine Learning},
  pages={13353--13409},
  year={2023},
  organization={PMLR}
}

@article{moradipari2020stage_conservative,
  title={Stage-wise conservative linear bandits},
  author={Moradipari, Ahmadreza and Thrampoulidis, Christos and Alizadeh, Mahnoosh},
  journal={Advances in neural information processing systems},
  volume={33},
  pages={11191--11201},
  year={2020}
}

@inproceedings{wu2016conservative,
  title={Conservative bandits},
  author={Wu, Yifan and Shariff, Roshan and Lattimore, Tor and Szepesv{\'a}ri, Csaba},
  booktitle={International Conference on Machine Learning},
  pages={1254--1262},
  year={2016},
  organization={PMLR}
}

@inproceedings{
bernasconi2024bandits_replenishable,
title={Bandits with Replenishable Knapsacks: the Best of both Worlds},
author={Martino Bernasconi and Matteo Castiglioni and Andrea Celli and Federico Fusco},
booktitle={The Twelfth International Conference on Learning Representations},
year={2024}
}

@inproceedings{tran2012knapsack,
  title={Knapsack based optimal policies for budget--limited multi--armed bandits},
  author={Tran-Thanh, Long and Chapman, Archie and Rogers, Alex and Jennings, Nicholas},
  booktitle={Proceedings of the AAAI Conference on Artificial Intelligence},
  volume={26},
  number={1},
  pages={1134--1140},
  year={2012}
}

@phdthesis{soare2015sequential,
  title={Sequential resource allocation in linear stochastic bandits},
  author={Soare, Marta},
  year={2015},
  school={Universit{\'e} Lille 1-Sciences et Technologies}
}

@article{villar2015multi,
  title={Multi-armed bandit models for the optimal design of clinical trials: benefits and challenges},
  author={Villar, Sof{\'\i}a S and Bowden, Jack and Wason, James},
  journal={Statistical science: a review journal of the Institute of Mathematical Statistics},
  volume={30},
  number={2},
  pages={199},
  year={2015},
  publisher={Europe PMC Funders}
}

@article{slivkins2013dynamic,
  title={Dynamic ad allocation: Bandits with budgets},
  author={Slivkins, Aleksandrs},
  journal={arXiv preprint arXiv:1306.0155},
  year={2013}
}

@article{intayoad2020reinforcement,
  title={Reinforcement learning based on contextual bandits for personalized online learning recommendation systems},
  author={Intayoad, Wacharawan and Kamyod, Chayapol and Temdee, Punnarumol},
  journal={Wireless Personal Communications},
  volume={115},
  number={4},
  pages={2917--2932},
  year={2020},
  publisher={Springer}
}

@article{badanidiyuru2018bandits,
  title={Bandits with knapsacks},
  author={Badanidiyuru, Ashwinkumar and Kleinberg, Robert and Slivkins, Aleksandrs},
  journal={Journal of the ACM (JACM)},
  volume={65},
  number={3},
  pages={1--55},
  year={2018},
  publisher={ACM New York, NY, USA}
}

@inproceedings{yu2016linear_bwk_recommendation_system,
  title={Linear submodular bandits with a knapsack constraint},
  author={Yu, Baosheng and Fang, Meng and Tao, Dacheng},
  booktitle={Proceedings of the AAAI Conference on Artificial Intelligence},
  volume={30},
  number={1},
  year={2016}
}

@article{kumar2022non_resource_alloc_bwk,
  title={Non-monotonic resource utilization in the bandits with knapsacks problem},
  author={Kumar, Raunak and Kleinberg, Robert},
  journal={Advances in Neural Information Processing Systems},
  volume={35},
  pages={19248--19259},
  year={2022}
}

@inproceedings{bwk_online_advertise,
  title={Stochastic bandits for multi-platform budget optimization in online advertising},
  author={Avadhanula, Vashist and Colini Baldeschi, Riccardo and Leonardi, Stefano and Sankararaman, Karthik Abinav and Schrijvers, Okke},
  booktitle={Proceedings of the Web Conference 2021},
  pages={2805--2817},
  year={2021}
}

@inproceedings{agrawal2014bandits,
  title={Bandits with concave rewards and convex knapsacks},
  author={Agrawal, Shipra and Devanur, Nikhil R},
  booktitle={Proceedings of the fifteenth ACM conference on Economics and computation},
  pages={989--1006},
  year={2014}
}

@article{Badanidiyuru2013BanditsWK,
  title={Bandits with Knapsacks},
  author={Ashwinkumar Badanidiyuru and Robert D. Kleinberg and Aleksandrs Slivkins},
  journal={2013 IEEE 54th Annual Symposium on Foundations of Computer Science},
  year={2013},
  pages={207-216}
}

@inproceedings{li2021symmetry_knap_1,
  title={The symmetry between arms and knapsacks: A primal-dual approach for bandits with knapsacks},
  author={Li, Xiaocheng and Sun, Chunlin and Ye, Yinyu},
  booktitle={International Conference on Machine Learning},
  pages={6483--6492},
  year={2021},
  organization={PMLR}
}

@article{flajolet2015logarithmic_knap_2,
  title={Logarithmic regret bounds for bandits with knapsacks},
  author={Flajolet, Arthur and Jaillet, Patrick},
  journal={arXiv preprint arXiv:1510.01800},
  year={2015}
}

@article{lai1985asymptotically,
  author  = {Tze Leung Lai and Herbert Robbins},
  title   = {Asymptotically Efficient Adaptive Allocation Rules},
  journal = {Advances in Applied Mathematics},
  volume  = {6},
  number  = {1},
  pages   = {4--22},
  year    = {1985}
}

@inproceedings{combes2015bandits,
  author    = {Richard Combes and Chong Jiang and R. Srikant},
  title     = {Bandits with Budgets: Regret Lower Bounds and Optimal Algorithms},
  booktitle = {Proceedings of the 2015 ACM SIGMETRICS International Conference on Measurement and Modeling of Computer Systems},
  pages     = {245--257},
  year      = {2015},
  publisher = {ACM},
  doi       = {10.1145/2745844.2745847}
}

\newpage

\appendix

\subsection{Proof of Lemma \ref{lem:omega_bounds}}\label{app:proof_lem_omega_bounds}

\begin{proof}
When the LP step is reached, the broad interval $[\varrho_i^L(t),\varrho_i^U(t)]$ excludes $c$ for every arm. Hence $|\Bar\rho_i(t)-c|>7a_i(t)$. On the good cost event, $|\Bar\rho_i(t)-\rho_i|\le a_i(t)$, which implies $\delta_i>6a_i(t)$. Therefore
\begin{align}
 |\Bar\rho_i(t)-c|-a_i(t)
 &\ge \delta_i-2a_i(t)\ge \frac{2}{3}\delta_i,\\
 |\Bar\rho_i(t)-c|-a_i(t)&\le \delta_i.
\end{align}
Taking minima gives $\frac{2}{3}\delta_{\min}\le\delta_{\min}^L(t)\le\delta_{\min}$. Since $x\mapsto x/(2+x-c)$ is increasing for $x\ge0$ and $2+\delta_{\min}-c\le3$, substituting these two bounds in \eqref{eq:omega_t} yields \eqref{eq:omega_bounds}.
\end{proof}

\subsection{Proof of Lemma \ref{lem:dual_decomposition}}\label{app:proof_lem_dual_decomposition}

\begin{proof}
The dual follows from the per-round normalization of \eqref{def:opt_lp}. The null-arm constraint gives $\nu^*\ge0$. Strong duality and rewards in $[0,1]$ give
\begin{align}
 0\le c\lambda^*\le\nu^*+c\lambda^*=r^*\le1,
\end{align}
so $\lambda^*\le1/c$.
For a canonical two-arm base, $\sum_q\pi_q^{\mathcal I}=1$ and
$\sum_q\pi_q^{\mathcal I}\rho_q=c$. Hence
\begin{align}
 \Delta_{\mathcal I}
 &=\nu^*+c\lambda^*-
   \sum_{q\in\mathcal I}\pi_q^{\mathcal I}\mu_q \\
 &=\sum_{q\in\mathcal I}\pi_q^{\mathcal I}
   (\nu^*+\lambda^*\rho_q-\mu_q)
 =\sum_{q\in\mathcal I}\pi_q^{\mathcal I}g_q.
\end{align}
The canonical weights are strictly positive and sum to one. Since $g_q\ge0$, this identity implies $\Delta_{\mathcal I}\le G_{\mathcal I}$, and $\Delta_{\mathcal I}=0$ if and only if $G_{\mathcal I}=0$.
For a single arm $\{i\}$,
\begin{align}
 \Delta_{\{i\}}
 =g_i+\lambda^*(c-\rho_i)\ge g_i,
\end{align}
and $G_{\{i\}}=\Delta_{\{i\}}$ by definition.

Conditional on $\mathcal F_{t-1}$, the selected support $\mathcal I_t$ and its within-base sampling probabilities are fixed, since they are determined from past observations and deterministic LP tie-breaking. Let $A_t$ denote the arm subsequently drawn from $\mathcal I_t$. By \eqref{eq:conditional_means_freshness}, conditional on $\mathcal F_{t-1}$ and $A_t$, the current reward and cost have means $\mu_{A_t}$ and $\rho_{A_t}$, respectively. Therefore the dual identity gives
\begin{align}
 r^*-\E[r(t)\mid\mathcal F_{t-1}]
 &=\E[g_{A_t}\mid\mathcal F_{t-1}]
 +\lambda^*\left(c-\E[c(t)\mid\mathcal F_{t-1}]\right).
\end{align}
For any mixture on a canonical selected support, the first conditional expectation is at most $G_{\mathcal I_t}$, proving \eqref{eq:dual_round_decomposition}. For a single-arm base the arm is pulled with probability one, so its expected regret is exactly $\Delta_{\mathcal I_t}=G_{\mathcal I_t}$.
\end{proof}

\subsection{Proof of Lemma \ref{lem:cost_classification}}\label{app:proof_lem_cost_classification}

\begin{proof}
We prove the three claims in the lemma separately.

\paragraph{Number of good-round classification pulls}
Fix a non-null arm \(i\in[K]\), and recall that
\begin{align}
\delta_i:=|\rho_i-c|,
\qquad
a_i(t):=\sqrt{\frac{1.5\log t}{N_i(t)}}.
\end{align}
When \(N_i(t)=0\), the algorithm sets
\(\varrho_i^L(t)=-\infty\) and
\(\varrho_i^U(t)=+\infty\), so arm \(i\) is necessarily cost-uncertain. This can cause at most one classification pull before the arm has been observed for the first time. We therefore consider below only classification pulls occurring when \(N_i(t)>0\).

If arm \(i\) is pulled by the classification rule on a good round \(t\), then its broad interval contains \(c\), so
\begin{align}
 |\Bar\rho_i(t)-c|\le 7a_i(t).
\end{align}
On \(\mathcal G_t\), \(|\Bar\rho_i(t)-\rho_i|\le a_i(t)\). Hence, for either \(\rho_i>c\) or \(\rho_i<c\),
\begin{align}
 \delta_i
 =|\rho_i-c|
 \le |\rho_i-\Bar\rho_i(t)|+|\Bar\rho_i(t)-c|
 \le 8a_i(t).
\end{align}
Therefore every good-round classification pull with \(N_i(t)>0\) satisfies
\begin{align}
 N_i(t)
 \le \frac{96\log t}{\delta_i^2}
 \le \frac{96\log T}{\delta_i^2}.
\end{align}
Because \(N_i(t)\) is the pre-round observation count and each pull increments it by one, there are at most \(96\log T/\delta_i^2\) such pulls after the first observation, plus at most one initial pull when \(N_i(t)=0\). Thus
\begin{align}
N_i^{{\rm cls},G}(T)
\le
\frac{96\log T}{\delta_i^2}+1,
\end{align}
which proves \eqref{eq:classification_pulls}.

\paragraph{Number of dedicated classification skips}
We next bound the number of skips generated by the dedicated classification cost tracker.
Consider any round on which the dedicated classification-skip condition is triggered. Immediately before that skip, let
\begin{align}
m&:=\text{number of classification pulls already executed},\\
s&:=\text{number of dedicated classification skips already executed}.
\end{align}
Since the classification cost tracker is updated only on these two types of rounds,
\begin{align}
N_p=m+s.
\end{align}
Moreover, every classification pull incurs a cost in \([0,1]\), while a dedicated skip incurs zero cost. Hence the cumulative cost incurred satisfies
\begin{align}
S_p\le m.
\label{eq:classification_account_cost_bound}
\end{align}

A dedicated skip occurs only if
\begin{align}
S_p+1>cN_p.
\end{align}
Using \(N_p=m+s\) and \eqref{eq:classification_account_cost_bound},
\begin{align}
c(m+s)
<
S_p+1
\le
m+1.
\end{align}
Therefore
\begin{align}
cs
<
(1-c)m+1,
\end{align}
and hence
\begin{align}
s
<
\frac{1-c}{c}m+\frac1c.
\end{align}
After performing the current skip, the number of dedicated skips becomes \(s+1\), so
\begin{align}
s+1
&<
\frac{1-c}{c}m+\frac1c+1\\
&\le
\frac{m}{c}+\frac1c+1,
\label{eq:cls_skip_intermediate}
\end{align}
where the last inequality uses \(1-c\le1\).

Now consider the last dedicated classification skip, if any, that occurs by time \(T\). At that time, the number \(m\) of classification pulls already executed is no larger than the total number \(M_{\rm cls}(T)\) of classification pulls through horizon \(T\). Applying \eqref{eq:cls_skip_intermediate} to this last skip gives
\begin{align}
S_{\rm cls}(T)
\le
\frac{M_{\rm cls}(T)}{c}
+\frac1c+1.
\end{align}
If no dedicated classification skip occurs, the same inequality is immediate. Thus \eqref{eq:classification_skips} holds pathwise.

\paragraph{Total number of classification pulls.}
Finally, decompose the total number of classification pulls according to whether the confidence event holds:
\begin{align}
M_{\rm cls}(T)
=
M_{\rm cls}^{G}(T)
+
M_{\rm cls}^{F}(T),
\end{align}
where \(M_{\rm cls}^{G}(T)\) and
\(M_{\rm cls}^{F}(T)\) denote classification pulls on rounds satisfying
\(\mathcal G_t\) and \(\mathcal G_t^c\), respectively.

Summing \eqref{eq:classification_pulls} over all non-null arms gives the pathwise bound
\begin{align}
M_{\rm cls}^{G}(T)
&=
\sum_{i=1}^K N_i^{{\rm cls},G}(T)\\
&\le
\sum_{i=1}^K
\left(
\frac{96\log T}{\delta_i^2}+1
\right).
\label{eq:good_classification_total}
\end{align}

On a confidence-failure round, the algorithm can execute at most one classification pull. Therefore
\begin{align}
M_{\rm cls}^{F}(T)
\le
\sum_{t=1}^T\ind{\mathcal G_t^c}
=
F_T.
\label{eq:bad_classification_total}
\end{align}
Combining \eqref{eq:good_classification_total} and
\eqref{eq:bad_classification_total} yields, pathwise,
\begin{align}
M_{\rm cls}(T)
&\le
96\sum_{i=1}^K
\frac{\log T}{\delta_i^2}
+K+F_T.
\end{align}
Taking expectations on both sides gives
\begin{align}
\E[M_{\rm cls}(T)]
\le
96\sum_{i=1}^K
\frac{\log T}{\delta_i^2}
+K+\E[F_T],
\end{align}
which is \eqref{eq:classification_total_pulls} and completes the proof.
\end{proof}

\subsection{Proof of Lemma \ref{lem:suboptimal_base}}\label{app:proof_lem_suboptimal_base}

\begin{proof}
We first fix a selected suboptimal two-arm base $\mathcal I=\{h,\ell\}$, with $\rho_h>c>\rho_\ell$, on a good round $t$. Define
\begin{align}
 D_{\mathcal I}&:=\rho_h-\rho_\ell=\delta_h+\delta_\ell\le1,
 &m_{\mathcal I}&:=\min\{\delta_h,\delta_\ell\},\\
 W_{\mathcal I}(t)&:=\delta_\ell\epsilon_h(t)+\delta_h\epsilon_\ell(t),
 &d_q(t)&:=\rho_q-\rho_q^L(t).
\end{align}
The weight of high-cost arm $h$ is $\pi_h=\delta_\ell/D_{\mathcal I}$, and the smallest weight is $m_{\mathcal I}/D_{\mathcal I}$. The conventions $\delta_{\bot}=c$ and $\epsilon_{\bot}(t)=d_{\bot}(t)=0$ apply if $\ell=\bot$.

On $\mathcal G_t$, $0\le d_q(t)\le2\epsilon_q(t)$. The LP step is reached only after the broad cost intervals exclude $c$, so $\delta_q>6a_q(t)$ and $\epsilon_q(t)=\sqrt2a_q(t)\le\delta_q/4$ for every non-null member. In particular, the optimistic cost bounds remain on the correct sides of $c$, and
\begin{align}
 \rho_h^L(t)-\rho_\ell^L(t)
 &=D_{\mathcal I}-d_h(t)+d_\ell(t)
 \ge D_{\mathcal I}/2.
\end{align}
Writing $\widetilde\pi_h(t)$ for the high-cost weight of the optimistic LP solution, we obtain
\begin{align}
 \widetilde\pi_h(t)
 &=\frac{\delta_\ell+d_\ell(t)}
 {D_{\mathcal I}-d_h(t)+d_\ell(t)},\\
 0\le\widetilde\pi_h(t)-\pi_h
 &=\frac{\delta_h d_\ell(t)+\delta_\ell d_h(t)}
 {D_{\mathcal I}(D_{\mathcal I}-d_h(t)+d_\ell(t))}
 \le\frac{4W_{\mathcal I}(t)}{D_{\mathcal I}^2}.
\end{align}
Since the selected optimistic value $\widetilde r_{\mathcal I}(t)$ is at least $r^*$, adding and subtracting $\sum_{q\in\mathcal I}\pi_q\mu_q^U(t)$ gives
\begin{align}
 \Delta_{\mathcal I}
 &\le\widetilde r_{\mathcal I}(t)-r_{\mathcal I}\\
 &=\sum_{q\in\mathcal I}\pi_q\bigl(\mu_q^U(t)-\mu_q\bigr)
 +(\widetilde\pi_h(t)-\pi_h)\bigl(\mu_h^U(t)-\mu_\ell^U(t)\bigr)\\
 &\le2\sum_{q\in\mathcal I}\pi_q\epsilon_q(t)
 +|\widetilde\pi_h(t)-\pi_h|\\
 &\le\left(\frac{2}{D_{\mathcal I}}+\frac{4}{D_{\mathcal I}^2}\right)
 W_{\mathcal I}(t).
 \label{eq:optimism_gap_condition}
\end{align}
Here $\mu_q^U(t)-\mu_q\le2\epsilon_q(t)$ on $\mathcal G_t$, and the clipped reward UCBs satisfy $|\mu_h^U(t)-\mu_\ell^U(t)|\le1$.

By \eqref{eq:base_gap_reduced_cost}, nonnegative reduced costs and the smallest canonical weight give
\begin{align}
 G_{\mathcal I}
 &\le\frac{D_{\mathcal I}}{m_{\mathcal I}}\Delta_{\mathcal I}\\
 &\le(2D_{\mathcal I}+4)\frac{W_{\mathcal I}(t)}{m_{\mathcal I}D_{\mathcal I}}\\
 &\le6\left(\frac{\epsilon_h(t)}{\delta_h}
            +\frac{\epsilon_\ell(t)}{\delta_\ell}\right)
 \le12\max_{q\in\mathcal I}\frac{\epsilon_q(t)}{\delta_q}.
 \label{eq:reduced_cost_optimism}
\end{align}
The third line uses $D_{\mathcal I}\le1$ and
$\delta_h\delta_\ell\le m_{\mathcal I}D_{\mathcal I}$. Since $G_{\mathcal I}>0$ and the null arm has zero confidence radius, some non-null $q\in\mathcal I$ satisfies
\begin{align}
 \frac{1}{\delta_q}\sqrt{\frac{3\log t}{N_q(t)}}
 \ge\frac{G_{\mathcal I}}{12},
\end{align}
which proves \eqref{eq:base_sample_threshold}. For a selected singleton $\{i\}$, optimism implies $G_{\{i\}}=\Delta_{\{i\}}\le2\epsilon_i(t)$; since $\delta_i\le1$, the same threshold is conservative. A suboptimal null singleton cannot be selected on a good round because its optimistic value is zero.

We next apply gap-layer counting to $G_{\mathcal I}$. Fix a non-null arm $i$ that belongs to at least one suboptimal base, and let
\begin{align}
 \sigma_{i,1}\ge\sigma_{i,2}\ge\cdots\ge\sigma_{i,m_i}=G_{\min,i}
\end{align}
be the distinct positive charges of bases containing $i$. Define
\begin{align}
 H_{i,j}:=\frac{432\log T}{\delta_i^2\sigma_{i,j}^2}.
\end{align}
A good round charged to $i$ with $G_{\mathcal I_t}\ge\sigma_{i,j}$ can occur only while $N_i(t)\le H_{i,j}$. Let $A_{i,j,t}$ indicate such a round, $Y_{i,t}$ indicate that $i$ is actually pulled, and $C_{i,j}:=\sum_t A_{i,j,t}$. The charge is determined before the within-base draw, and $A_{i,j,t}$ is $\mathcal F_{t-1}$-measurable. On a charged round, Lemma~\ref{lem:omega_bounds} gives sampling probability at least $\omega_{\min}=2\delta_{\min}/9$; the probability is one for a singleton. Hence
\begin{align}
 \omega_{\min}\E[C_{i,j}]
 &\le\E\!\left[\sum_t A_{i,j,t}Y_{i,t}\right]
 \le H_{i,j}.
\end{align}
The final inequality is pathwise: before any LP round, every non-null arm has already been sampled at least once, and each charged draw of $i$ increments $N_i$ by one. Once $N_i>H_{i,j}$, no good round with charge at least $\sigma_{i,j}$ can be assigned to $i$. Observations from other bases or classification pulls only increase $N_i$. Therefore
\begin{align}
 \E[C_{i,j}]\le\frac{H_{i,j}}{\omega_{\min}}.
 \label{eq:cumulative_gap_charges}
\end{align}
Summation by parts over the charge layers gives
\begin{align}
 R_{{\rm str},i}(T)
 &\le\frac{432\log T}{\omega_{\min}\delta_i^2}
 \left[
 \frac{1}{\sigma_{i,m_i}}
 +\sum_{j=1}^{m_i-1}
 \frac{\sigma_{i,j}-\sigma_{i,j+1}}{\sigma_{i,j}^2}
 \right]\\
 &\le\frac{864\log T}{\omega_{\min}\delta_i^2G_{\min,i}}
 =\frac{3888\log T}{\delta_{\min}\delta_i^2G_{\min,i}}.
\end{align}
The second inequality follows from
\begin{align}
 \frac{\sigma_{i,j}-\sigma_{i,j+1}}{\sigma_{i,j}^2}
 \le\int_{\sigma_{i,j+1}}^{\sigma_{i,j}}\frac{dx}{x^2}.
\end{align}
Since $G_{\mathcal I}\ge\Delta_{\mathcal I}$ and the positive-charge and suboptimal bases coincide, $G_{\min,i}\ge\Delta_{\min,i}$, proving \eqref{eq:arm_charged_base_regret}. If no suboptimal base contains $i$, it receives no positive charge and the stated bound is zero. Finally, Lemma~\ref{lem:dual_decomposition} bounds the actual reduced-cost loss on a two-arm LP round by $G_{\mathcal I_t}$, while singleton regret equals $G_{\mathcal I_t}$. Both are therefore covered by this single charging argument.
\end{proof}

\subsection{Proof of Lemma \ref{lem:skip_underbudget}}\label{app:proof_lem_skip_underbudget}

\begin{proof}
Define the cumulative slack contributed by good singleton LP rounds by
\begin{align}
 U_t:=
 \sum_{\substack{s\le t:\,\mathcal G_s,\ |\mathcal I_s|=1}}
 (c-c(s)).
\end{align}
On a good singleton LP round the selected arm has true mean cost below $c$, so
\begin{align}
 \E[c-c(s)\mid\mathcal F_{s-1}]
 =c-\rho_{i(s)}\ge\delta_{\min}.
\end{align}
Since $c-c(s)\in[-1,1]$, Hoeffding's lemma implies that
$\exp(-\delta_{\min}U_t)$ is a nonnegative supermartingale on the sequence of good singleton LP rounds. Hence, by Ville's inequality,
\begin{align}
 \Pr\!\left(\inf_{s\le T}U_s<-B_T\right)\le T^{-4},
 \qquad
 B_T:=\frac{4\log T}{\delta_{\min}}.
\label{eq:singleton_slack_event}
\end{align}
Let $\mathcal H_T:=\{U_s\ge-B_T,\ \forall s\le T\}$.
Let $L_t:=ct-S_c(t)$ for the actual global slack. The classification cost tracker satisfies
$cN_p(t)-S_p(t)\ge0$ pathwise by construction. Because $N_{\rm gskip}^{\rm G}$ counts only rounds on which the global-skip branch is actually executed, dedicated classification cost tracker skips are represented only through this nonnegative classification cost tracker slack and are not counted again in $Q_t$. Good singleton LP rounds contribute $U_t$, good two-arm LP rounds and actual good global skips contribute exactly $Q_t$, and every remaining confidence-failure round can reduce slack by at most one. Therefore
\begin{align}
 L_t\ge Q_t+U_t-F_t.
\end{align}
On $\mathcal H_T$,
\begin{align}
 L_t\ge \widetilde Q_t-B_T.
\label{eq:slack_joint_Q}
\end{align}

Let
\begin{align}
 R_t:=\frac{\log t}{\omega^2(t)},
 \qquad
 \overline R_T:=\frac{81\log T}{4\delta_{\min}^2}.
\end{align}
Lemma~\ref{lem:omega_bounds} gives $R_t\le\overline R_T$. Thus, on a good two-arm LP round,
\begin{align}
 b(t)=c+L_{t-1}-R_t
 \ge c+\widetilde Q_{t-1}-B_T-\overline R_T.
\end{align}
Hence, whenever
\begin{align}
 \widetilde Q_{t-1}>a_T:=\overline R_T+B_T+1-c,
\label{eq:joint_threshold}
\end{align}
we have $b(t)>1$, and therefore SUAK uses $p(t)=1-\omega(t)$ for the high-cost arm. If $j(t)$ and $k(t)$ are the high- and low-cost members of the selected base, then the high-cost arm is selected with probability $1-\omega(t)$ and the low-cost arm with probability $\omega(t)$. Hence
\begin{align}
 m_t-c
 &= (1-\omega(t))\delta_{j(t)}-\omega(t)\delta_{k(t)}\\
 &\ge (1-\omega_{\max})\delta_{\min}-\omega_{\max}c\\
 &= \frac{2(1-c)\delta_{\min}}{2+\delta_{\min}-c}\\
 &\ge \frac23\delta_{\min}^2=:d.
\label{eq:joint_negative_drift}
\end{align}
Here the first inequality uses $\delta_{j(t)}\ge\delta_{\min}$, $\delta_{k(t)}=c-\rho_{k(t)}\le c$, and $\omega(t)\le\omega_{\max}=\delta_{\min}/(2+\delta_{\min}-c)$. Moreover, because a two-arm base contains a high-cost arm with $\rho_{j(t)}>c$ and all costs lie in $[0,1]$, we have
\begin{align}
 \delta_{\min}\le\delta_{j(t)}=\rho_{j(t)}-c\le1-c.
\end{align}
Consequently $1-c\ge\delta_{\min}$ and
$2+\delta_{\min}-c\le3$, which gives the final inequality in \eqref{eq:joint_negative_drift}. Thus, whenever the upper slack barrier is crossed, the good two-arm increments $Z_t=c-c(t)$ have conditional mean at most $-(2/3)\delta_{\min}^2$.

A global skip cannot occur while $\widetilde Q_{t-1}>a_T$: indeed, \eqref{eq:slack_joint_Q} gives
$L_{t-1}>\overline R_T+1-c\ge1-c$. Thus, above $a_T$, the only good-round increments of $\widetilde Q_t$ are two-arm LP increments $Z_t=c-c(t)\in[-1,1]$, whose conditional mean is at most $-d$, while a confidence-failure step decreases $\widetilde Q_t$ by one. Hoeffding's lemma gives
\begin{align}
 \E[e^{dZ_t}\mid\mathcal F_{t-1}]
 \le \exp\!\left(d\E[Z_t\mid\mathcal F_{t-1}]+\frac{d^2}{2}\right)
 \le e^{-d^2/2}<1.
\end{align}
Let
\begin{align}
 \tau_H:=\inf\{t\ge1:U_t<-B_T\}.
\end{align}
During an excursion of $\widetilde Q_t$ above $a_T$, stopped at $\tau_H$, the drift condition \eqref{eq:joint_negative_drift} holds on every good two-arm LP step, while a confidence-failure step decreases $\widetilde Q_t$ by one. Hence the stopped exponential process $e^{d\widetilde Q_{t\wedge\tau_H}}$ is a nonnegative supermartingale until the excursion returns to $(-\infty,a_T]$. Every increment of $Q_t$ is at most one, so the overshoot at the start of an excursion is at most one. Optional stopping and a union bound over at most $T$ excursions yield
\begin{align}
 \Pr\!\left(
 \sup_{t\le T}\widetilde Q_t>a_T+1+x,\ \tau_H>T
 \right)
 \le T e^{-dx}.
\end{align}
Since $\{\tau_H>T\}=\mathcal H_T$, this is exactly the desired bound on the good singleton-slack event.
Taking $x=4\log T/d$ gives a bound $T^{-3}$. Since $\widetilde Q_T\le T$, and $\Pr(\mathcal H_T^c)\le T^{-4}$,
\begin{align}
 \E[\widetilde Q_T]
 &\le \overline R_T+B_T+2+\frac{4\log T}{d}+T^{-2}+T^{-3}\\
 &\le \frac{41\log T}{\delta_{\min}^2},
 \qquad T\ge3.
\label{eq:joint_Q_bound}
\end{align}
Here we used $B_T\le4\log T/\delta_{\min}^2$ and $4/d=6/\delta_{\min}^2$.

Finally, membership in $\mathcal T_2^{\rm G}(T)$, the selected support, and the within-base sampling probabilities are $\mathcal F_{t-1}$-measurable before observing $c(t)$; the realized within-base draw $i(t)$ is not predictable. A global skip has deterministic cost zero. Conditioning first on $\mathcal F_{t-1}$ and the realized action and then using \eqref{eq:conditional_means_freshness} gives $\E[c(t)\mid\mathcal F_{t-1},i(t)]=\rho_{i(t)}$ on every non-null pull. Hence
\begin{align}
 \E[Q_T]
 =\E\!\left[
 \sum_{t\in\mathcal T_2^{\rm G}(T)}(c-\rho_{i(t)})
 +cN_{\rm gskip}^{\rm G}(T)
 \right].
\end{align}
Since $Q_T=\widetilde Q_T+F_T$, \eqref{eq:joint_Q_bound} proves \eqref{eq:joint_cost_tracking}--\eqref{eq:joint_cost_tracking_mean}.
\end{proof}

\subsection{Proof of Lemma \ref{lem:global_skip_count}}\label{app:proof_lem_global_skip_count}

\begin{proof}
Assume $\nu^*>0$. The null-arm dual constraint has reduced-cost gap $g_{\bot}=\nu^*>0$. Complementary slackness therefore gives $\pi^*_{\bot}=0$. Since $\pi^*$ is feasible, supported only on non-null arms, and satisfies $\sum_{i=1}^K\rho_i\pi_i^*\le c$, at least one non-null arm in its support must have $\rho_i\le c$. Assumption~\ref{ass:positive_cost_gap} rules out equality, so in fact $\rho_i<c$ for some non-null arm. Hence
\begin{align}
 \delta_{\min}\le c-\rho_i\le c.
\label{eq:delta_min_le_c_when_nu_pos}
\end{align}
This observation is exactly what is needed when a selected two-arm base uses the null arm as its low-cost member.

Let $L_t=ct-S_c(t)$ and define the lower reserve
\begin{align}
 \underline R_t:=\frac{\log t}{\omega_{\max}^2}.
\end{align}
Since $\omega(t)\le\omega_{\max}$, $R_t=\log t/\omega^2(t)\ge\underline R_t$. On a good ordinary LP round satisfying
$L_{t-1}\le\underline R_t-c$, we have $b(t)=c+L_{t-1}-R_t\le0$. Thus, for a two-arm base with high-cost member $h$ and low-cost member $\ell$, SUAK chooses $h$ with probability $\omega(t)$, and
\begin{align}
 \E[L_t-L_{t-1}\mid\mathcal F_{t-1}]
 &=\delta_{\ell}-\omega(t)(\delta_h+\delta_{\ell})\\
 &\ge(1-\omega_{\max})\delta_\ell-\omega_{\max}(1-c)\\
 &\ge(1-\omega_{\max})\delta_{\min}-\omega_{\max}(1-c)\\
 &=\frac{\delta_{\min}}{2+\delta_{\min}-c}=\eta.
\label{eq:protective_slack_drift}
\end{align}
The first inequality uses $\omega(t)\le\omega_{\max}$ and the cost-separation bound $\delta_h+\delta_\ell=\rho_h-\rho_\ell\le1-c+\delta_\ell$. For the second inequality, $\delta_\ell\ge\delta_{\min}$ for a non-null low-cost member, while if $\ell=\bot$ then $\delta_\ell=c\ge\delta_{\min}$ by \eqref{eq:delta_min_le_c_when_nu_pos}. For a good singleton LP round the same conditional drift is at least $\delta_{\min}\ge\eta$. We will treat the case in which no high-cost arm exists separately below; in that case every good ordinary LP round is a singleton with positive slack drift at every slack level. An actual global skip, i.e. a round with $J_t^{\rm gskip}=1$, increases $L_t$ deterministically by $c$. A dedicated classification cost tracker skip also has zero cost and therefore increases $L_t$ by $c$; such rounds can only increase slack and decrease the potential $e^{-\eta L_t}$, so they may be ignored in the protective-drift upper bound on global skips.

Partition time into dyadic epochs $E_m=\{2^m,\ldots,2^{m+1}-1\}$ and define
\begin{align}
 H_m:=\frac{m\log2}{\omega_{\max}^2}-c.
\end{align}
Split the horizon at every classification-pull round and every confidence-failure round. Each such interrupting round is excluded from both neighboring intervals; the next interval starts from the post-action history of that round. There are at most $1+M_{\rm cls}(T)+F_T$ intervals. The decision to exclude round $t$ is $\mathcal F_{t-1}$-measurable. Thus an interval ending because of an interruption in round $t$ stops at the state $L_{t-1}$, before the interrupting action is executed. This convention also applies to all excursions below.

Suppose first that a high-cost arm exists, so $\delta_{\min}\le1-c$. For $t\in E_m$, use $H_m$ as the threshold for the pre-action slack $L_{t-1}$. A low-slack excursion begins when this slack is at most the current threshold and ends upon exit above the threshold, at the end of the interval, or at the horizon. We use the following boundary convention throughout. If a dyadic boundary $t=2^{m+1}$ is reached while an excursion is active, then after updating the threshold from $H_m$ to $H_{m+1}$ the same excursion continues whenever $L_{t-1}\le H_{m+1}$; no new entrance time is declared and the excursion is not counted again. A threshold-induced entrance at that boundary is declared only when no excursion was active immediately before the boundary and
$H_m<L_{t-1}\le H_{m+1}$. Thus epoch changes never restart an already active excursion. Dedicated classification cost tracker skips are retained as nonnegative slack increments.

The initial-epoch contribution is explicit. Since $\omega_{\max}<1/2$,
$\omega_{\max}^2<1/4<\log2$. Thus, for every $m\ge1$,
$m\log2>\omega_{\max}^2$, which is equivalent to $H_m>1-c$.
Hence the only epoch for which $H_m\le1-c$ is $E_0=\{1\}$, and round~$1$ contributes at most one global skip over the entire horizon. We isolate that round completely from the dyadic entrance count. The analysis for $m\ge1$ begins at round~$2$ from the post-round-one state $L_1$. If the restart interval containing round~$2$ satisfies $L_1\le H_1$, we declare the state $L_1$ to be that interval's initial low-slack excursion and assign it the trivial hitting-probability bound one; if $L_1>H_1$, no such initial excursion is present. In particular, the boundary $E_0\to E_1$ is absorbed by this initial-excursion convention and is not counted as a threshold-induced entrance. The threshold-induced entrance sum therefore starts with $m=1$, corresponding to subsequent dyadic boundaries.

For the excursion calculation below we consequently take $m\ge1$, so $H_m>1-c$. While an excursion consists of good ordinary LP rounds and global skips, \eqref{eq:protective_slack_drift} holds because, for every $t\in E_m$, $H_m\le \underline R_t-c$. For an ordinary increment $X_t=L_t-L_{t-1}\in[-1,1]$, Hoeffding's lemma gives
\begin{align}
 \E[e^{-\eta X_t}\mid\mathcal F_{t-1}]
 \le \exp\!\left(-\eta^2+\frac{\eta^2}{2}\right)<1,
\end{align}
where the looser range-two Hoeffding constant is used. Hence $e^{-\eta L_t}$ is a supermartingale during each low-slack excursion; a global skip also decreases this potential deterministically.

If an excursion is initiated during epoch $m$ by an ordinary pull crossing from above $H_m$ to at most $H_m$, then the post-pull slack is at least $H_m-(1-c)$. Optional stopping therefore gives a probability at most
\begin{align}
 \exp\!\left(-\eta[H_m-2(1-c)]\right)
\end{align}
of reaching the unsafe set $[0,1-c)$ before exiting the low-slack region. Moreover,
\begin{align}
 \frac{\eta}{\omega_{\max}^2}
 =\frac1{\omega_{\max}}
 =\frac{2+\delta_{\min}-c}{\delta_{\min}}>2.
\end{align}
Using $\eta=\omega_{\max}<1/2$ and $2-c<2$, we obtain the explicit bound
\begin{align}
 \exp\!\left(-\eta[H_m-2(1-c)]\right)
 &=\exp\!\left(-\frac{m\log2}{\omega_{\max}}+\eta(2-c)\right)\\
 &\le e\,4^{-m}.
\label{eq:ordinary_entrance_hit_explicit}
\end{align}
There are at most $2^m$ ordinary-pull entrances in epoch $m$.

For every subsequent boundary $E_m\to E_{m+1}$ with $m\ge1$, the boundary convention above implies that a new threshold-induced excursion can occur only if no excursion is already active just before the boundary and the increase from $H_m$ to $H_{m+1}$ newly places the current slack below the new threshold, i.e., $H_m<L_{t-1}\le H_{m+1}$. In that case the slack at the boundary is strictly larger than $H_m$. Starting from this value, the same exponential-supermartingale argument gives
\begin{align}
 \Pr(\text{hit }[0,1-c)\text{ before exit})
 &\le \exp\!\left(-\eta[H_m-(1-c)]\right)\\
 &=\exp\!\left(-\frac{m\log2}{\omega_{\max}}+\eta\right)\\
 &\le e^{1/2}4^{-m}
 \le e\,4^{-m}.
\label{eq:threshold_entrance_hit_explicit}
\end{align}
There is at most one such threshold-induced entrance per epoch. By construction, an excursion already active at an epoch boundary retains its original entrance time and is not counted again. At the beginning of each interval created by a classification/failure restart, we likewise allow at most one initial low-slack excursion with the trivial hitting-probability bound one. This same initial-excursion convention is what absorbs the post-round-one state at the $E_0\to E_1$ boundary. The explicit sum of these hitting probabilities is evaluated below. To convert these hitting-probability bounds into a bound on the number of global skips, we first bound the expected number of skips generated by an excursion conditional on its reaching the unsafe region \([0,1-c)\). We then multiply this conditional bound by the corresponding hitting probability and sum over all possible excursion entrances.

We next bound skips conditionally from any state inside an excursion. Set $V_t:=e^{-\eta L_t}$; pathwise feasibility gives $0<V_t\le1$. On an ordinary good LP action inside the low-slack region, the preceding calculation gives
\begin{align}
 \E[V_t-V_{t-1}\mid\mathcal F_{t-1}]\le0.
\end{align}
A dedicated classification cost tracker skip also decreases $V_t$. On an actual global skip, $L_{t-1}<1-c\le1$, so $V_{t-1}\ge e^{-1}$. Since $\eta c\le1$ and $1-e^{-\eta c}\ge\eta c/2$,
\begin{align}
 V_t-V_{t-1}
 &=V_{t-1}(e^{-\eta c}-1)
 \le-\frac{\eta c}{2e}.
\end{align}
Let $\sigma_e$ and $\kappa_e$ denote the entrance state and terminal state of an excursion, indexed by completed rounds, and let $I_{e,t}^{\rm skip}$ indicate a global skip in that excursion. By the pre-action stopping convention, every included action obeys the conditional drift inequality. Applying it after any stopping time $\theta$ with $\sigma_e\le\theta\le\kappa_e$, summing up to $\kappa_e\le T$, and conditioning on $\mathcal F_\theta$ gives
\begin{align}
 \E[V_{\kappa_e}\mid\mathcal F_\theta]-V_\theta
 &\le-\frac{\eta c}{2e}
 \E\!\left[\sum_{t=\theta+1}^{\kappa_e}I_{e,t}^{\rm skip}
 \,\middle|\,\mathcal F_\theta\right].
\end{align}
Using $V_{\kappa_e}\ge0$ and $V_\theta\le1$, we conclude that
\begin{align}
 \E\!\left[\sum_{t=\theta+1}^{\kappa_e}I_{e,t}^{\rm skip}
 \,\middle|\,\mathcal F_\theta\right]
 \le\frac{2e}{\eta c}.
 \label{eq:skips_per_excursion}
\end{align}

Let $\tau_e$ be the first state time in $[\sigma_e,\kappa_e]$ at which $L_{\tau_e}<1-c$, setting $\tau_e=\kappa_e$ if the unsafe region is not reached earlier. No global skip can occur before this first unsafe state; if $\tau_e=\kappa_e$, there are no global skips in the excursion. Applying \eqref{eq:skips_per_excursion} at $\theta=\tau_e$ and then the tower property therefore yields the conditional bound
\begin{align}
 \E[N_e^{\rm skip}\mid\mathcal F_{\sigma_e}]
 &\le\frac{2e}{\eta c}\,
 \Pr(\tau_e<\kappa_e\mid\mathcal F_{\sigma_e}).
 \label{eq:conditional_excursion_skip_bound}
\end{align}
The hitting estimates above are conditional on the entrance history and uniform over that history. Summing \eqref{eq:conditional_excursion_skip_bound} over entrances in a restart interval now gives an explicit constant. Since the excursion calculation starts at $m=1$,
\begin{align}
 \sum_{m=1}^{\infty}2^m e4^{-m}=e,
 \qquad
 \sum_{m=1}^{\infty}e4^{-m}=\frac e3.
\end{align}
Adding the interval's possible initial excursion, whose hitting probability is bounded by one, the total hitting-probability mass within any restart interval is at most
\begin{align}
 1+e+\frac e3=1+\frac{4e}{3}.
\end{align}
Therefore, by \eqref{eq:conditional_excursion_skip_bound}, conditional on the starting history of any restart interval, its expected number of global skips is at most
\begin{align}
 \frac{2e}{c\eta}\left(1+\frac{4e}{3}\right).
\label{eq:explicit_skips_per_restart_interval}
\end{align}

If no high-cost arm exists, every good ordinary LP action has mean slack increment at least $\delta_{\min}\ge\eta$ at every slack level. Hence the same potential drift and conditional skip-count calculation apply directly to the entire interval, with no threshold or excursion decomposition. Its conditional expected skip count is at most $2e/(c\eta)$, which is smaller than \eqref{eq:explicit_skips_per_restart_interval}.

Finally, condition successively on the histories at the starts of the intervals. The conditional bounds are uniform, and the number of intervals is at most $1+M_{\rm cls}(T)+F_T$. The isolated epoch $E_0=\{1\}$ contributes at most one additional global skip. Since $c\eta<1$ and $1+M_{\rm cls}(T)+F_T\ge1$,
\begin{align}
 \E[N_{\rm gskip}^{\rm G}(T)]
 &\le
 1+\frac{2e}{c\eta}\left(1+\frac{4e}{3}\right)
 \left(1+\E[M_{\rm cls}(T)]+\E[F_T]\right)\\
 &\le
 \frac{1+2e(1+4e/3)}{c\eta}
 \left(1+\E[M_{\rm cls}(T)]+\E[F_T]\right)\\
 &\le
 \frac{27}{c\eta}
 \left(1+\E[M_{\rm cls}(T)]+\E[F_T]\right),
\end{align}
where $1+2e(1+4e/3)<27$. This proves \eqref{eq:global_skip_count}. The simplified form follows from Lemma~\ref{lem:cost_classification} and $\eta\ge\delta_{\min}/3$.
\end{proof}

\subsection{Proof of Theorem \ref{thm:bwak_lower_bound}}\label{app:proof_thm_bwak_lower_bound}

\begin{proof}
Because $\rho_0<c$ and costs are deterministic, pulling any non-null arm in every round satisfies the anytime constraint pathwise:
\begin{align}
 \sum_{s=1}^{t} c(s)=t\rho_0<ct,
 \qquad \forall t.
\end{align}
Hence the null arm is never needed for feasibility and the budget constraint of the LP relaxation is slack. The optimal LP policy therefore places all mass on the unique arm with largest reward mean $\mu_*$, and $r^*=\mu_*$. To avoid confusion with the pre-round count $N_i(t)$ used elsewhere in the paper, define the horizon pull count
\begin{align}
 \widetilde N_i(T):=\sum_{t=1}^{T}\ind{i(t)=i},
 \qquad i\in\mathcal A.
\label{eq:bwak_horizon_count}
\end{align}
For any policy, the exact regret decomposition on this subclass is
\begin{align}
 R_T
 &=T\mu_*-\E\!\left[\sum_{t=1}^{T}r(t)\right]\nonumber\\
 &=\sum_{\substack{j\in[K]\\ j\neq i^*}}
 (\mu_*-\mu_j)\,\E[\widetilde N_j(T)]
 +\mu_*\,\E[\widetilde N_{\bot}(T)]\nonumber\\
 &\ge \sum_{j:\Delta_j>0}\Delta_j\,\E[\widetilde N_j(T)].
\label{eq:bwak_lower_regret_decomp}
\end{align}
Thus, after discarding the nonnegative regret due to null-arm pulls, the classical Lai--Robbins change-of-measure lower-bound argument \citep{lai1985asymptotically} applies to the non-null arms.

Fix a suboptimal arm $i$. Consider an alternative instance $\theta'$ that changes only arm $i$ and increases its Bernoulli reward mean from $\mu_i$ to some $\mu_i'>\mu_*$, while leaving all costs and all other reward distributions unchanged. Both instances belong to $\Theta_+$ because $\delta_{\min}=c-\rho_0>0$ in each. Let $\theta$ denote the original instance. Under $\theta$, arm $i$ is suboptimal, whereas under $\theta'$ it is the unique optimal arm. Since the two instances differ only in observations from arm $i$, the chain rule for relative entropy gives
\begin{align}
 D_{\rm KL}(\mathbb P_{\theta}^{T}\Vert\mathbb P_{\theta'}^{T})
 =\E_{\theta}[\widetilde N_i(T)]\,\operatorname{kl}(\mu_i,\mu_i').
\label{eq:bwak_history_kl}
\end{align}
To lower-bound this information, let $A_T:=\{\widetilde N_i(T)<T/2\}$. Uniform efficiency under $\theta$ and \eqref{eq:bwak_lower_regret_decomp} imply $\E_{\theta}[\widetilde N_i(T)]=o(T^a)$ for every $a>0$, and therefore $\mathbb P_{\theta}(A_T)\to1$. Under $\theta'$, every action different from arm $i$---including the null arm---has a strictly positive reward gap. Hence uniform efficiency under $\theta'$ implies $\E_{\theta'}[T-\widetilde N_i(T)]=o(T^a)$ for every $a>0$. Consequently, for every fixed $a>0$,
\begin{align}
 \mathbb P_{\theta'}(A_T)
 &\le \frac{2\E_{\theta'}[T-\widetilde N_i(T)]}{T}
 =o(T^{a-1}).
\label{eq:bwak_alt_event_prob}
\end{align}
Then,
\begin{align}
 D_{\rm KL}(\mathbb P_{\theta}^{T}\Vert\mathbb P_{\theta'}^{T})
 \ge
 \operatorname{kl}(\mathbb P_{\theta}(A_T),\mathbb P_{\theta'}(A_T)).
\end{align}
Since the first probability tends to one and the second is $o(T^{a-1})$, the right-hand side is at least $(1-a-o(1))\log T$. Combining this with \eqref{eq:bwak_history_kl}, dividing by $\log T$, and then letting $a\downarrow0$ yields
\begin{align}
 \liminf_{T\to\infty}
 \frac{\E_{\theta}[\widetilde N_i(T)]}{\log T}
 \ge
 \frac{1}{\operatorname{kl}(\mu_i,\mu_i')}.
\label{eq:bwak_lower_sampling_alt}
\end{align}
Letting $\mu_i'\downarrow\mu_*$ and using continuity of $\operatorname{kl}(\mu_i,\cdot)$ gives
\begin{align}
 \liminf_{T\to\infty}
 \frac{\E[\widetilde N_i(T)]}{\log T}
 \ge
 \frac{1}{\operatorname{kl}(\mu_i,\mu_*)}.
\label{eq:bwak_lower_sampling}
\end{align}
Multiplying by $\Delta_i$, summing over suboptimal arms, and applying \eqref{eq:bwak_lower_regret_decomp} proves \eqref{eq:bwak_lai_robbins}.

For the symmetric consequence, choose one arm with mean $\mu_*=1/2+\Delta$ and the remaining $K-1$ arms with mean $1/2$, where $0<\Delta\le1/4$. Then
\begin{align}
 \operatorname{kl}\!\left(\frac12,\frac12+\Delta\right)
 =-\frac12\log(1-4\Delta^2)
 =\Theta(\Delta^2).
\end{align}
Therefore each of the $K-1$ suboptimal arms contributes $\Theta((\log T)/\Delta)$ to the lower bound. Since $K-1=\Theta(K)$ for $K\ge2$, this yields \eqref{eq:bwak_symmetric_lower}.
\end{proof}

\end{document}